\documentclass[lettersize,journal]{IEEEtran}
\usepackage{amsmath,amsfonts}
\usepackage{algorithm}
\usepackage{algpseudocode}
\usepackage{array}
\usepackage{bm} 
\usepackage{booktabs} 
\usepackage{textcomp}
\usepackage{threeparttable}
\usepackage{stfloats}
\usepackage{url}
\usepackage{verbatim}
\usepackage{graphicx}
\usepackage{subfigure}
\usepackage{siunitx}
\usepackage{tabularx}
\usepackage{threeparttable}
\usepackage{multicol}
\usepackage{cite}

\begin{document}

\title{GBC: Generalized Behavior-Cloning Framework for Whole-Body Humanoid Imitation}

\author{Yifei Yao,
        Chengyuan Luo,
        Jiaheng Du,
        Wentao He,
        and Jun-Guo Lu\IEEEauthorrefmark{1}
    \thanks{All authors are with SJTU MVASL Lab, Department of Automation, Shanghai Jiao Tong University, Shanghai, China}
    \thanks{The authors' emails are (in order):\\ \{godchaser, luo\_cy, jiahengdu, goodmorning\_hwt, jglu\}@sjtu.edu.cn}
    \thanks{\IEEEauthorrefmark{1}J.-G. Lu is the corresponding author.}
}



\maketitle

\begin{abstract}
The creation of human-like humanoid robots is hindered by a fundamental fragmentation: data processing and learning algorithms are rarely universal across different robot morphologies. This paper introduces the Generalized Behavior Cloning (GBC) framework, a comprehensive and unified solution designed to solve this end-to-end challenge. GBC establishes a complete pathway from human motion to robot action through three synergistic innovations. First, an adaptive data pipeline leverages a differentiable IK network to automatically retarget any human MoCap data to any humanoid. Building on this foundation, our novel DAgger-MMPPO algorithm with its MMTransformer architecture learns robust, high-fidelity imitation policies. To complete the ecosystem, the entire framework is delivered as an efficient, open-source platform based on Isaac Lab, empowering the community to deploy the full workflow via simple configuration scripts. We validate the power and generality of GBC by training policies on multiple heterogeneous humanoids, demonstrating excellent performance and transfer to novel motions. This work establishes the first practical and unified pathway for creating truly generalized humanoid controllers.

\end{abstract}

\begin{IEEEkeywords}
Humanoid Robots, Reinforcement Learning, Imitation Learning, Whole-Body Control, Behavior Cloning, Transformers
\end{IEEEkeywords}

\section{Introduction}
Humanoid robotics, positioned at the cutting edge of artificial intelligence and robotic control, has witnessed significant research advancements in recent years. Their multi-degree-of-freedom (DoF) structures, kinematically similar to humans, enable them to perform human-like tasks, thereby showcasing substantial application potential across numerous service and industrial sectors \cite{saeedvand2019comprehensive}. However, designing a general control policy that enables them to move as naturally as humans is challenging due to their high-dimensional action spaces arising from numerous DoFs, complex nonlinear dynamics, and intricate mechanical interaction conditions \cite{sugihara2020survey, apgar2018fast}. Although considerable research has focused on achieving humanoid control through traditional methods such as Model Predictive Control (MPC)\cite{garcia1989model} and Whole-Body Control (WBC), or via pure Reinforcement Learning (RL) approaches \cite{feng2014optimization, chen2025squat, kuindersma2016optimization, wang2025walking,karoly2020deep, gu2024humanoid}, these often necessitate complex mathematical modeling, extensive stability analyses, task-specific motion solvers, or highly tailored reward function designs. Consequently, they encounter bottlenecks in tasks requiring broad generalization and high-fidelity human-like performance, struggling to cover the diverse scenarios that robots may need to address \cite{moro2019whole}.

To enable humanoids to better mimic human behaviors and execute tasks, research has increasingly focused on Imitation Learning (IL) \cite{hussein2017imitation} and robot teleoperation \cite{darvish2023teleoperation}. These approaches are typically data-driven, leveraging human demonstrations to allow robots to learn corresponding control policies either offline or online. Works based on teleoperation for demonstration \cite{fu2024mobile, fu2024humanplus, he2024learning} can achieve high-fidelity pose tracking and can even assist developers in recording high-quality motion data via teleoperation for subsequent training phases. Offline learning, commonly employing Behavior Cloning (BC), learns distributional features from existing expert actions and has demonstrated strong performance on numerous tasks \cite{peng2021amp, liang2024constrained, he2024omnih2o}. Notably, the concept of training general-purpose controllers via Behavior Cloning by using large-scale demonstration data resonates with the recent successes of Vision-Language-Action (VLA) models in manipulation tasks \cite{kim2024openvla, xu2024humanvla}, suggesting a promising direction for achieving human-like motion control in humanoids. However, achieving a truly Generalized Behavior Cloning (GBC) framework still faces critical challenges, primarily due to the wide variety of humanoid robots currently in use (e.g., Unitree H1, Atlas, Optimus), which exhibit significant differences in their morphologies, kinematic and dynamic characteristics (i.e., heterogeneous morphologies), and degrees of control freedom \cite{sheng2025comprehensive}.

First, there is the challenge of acquiring high-quality, generalizable robot demonstration data. Due to inherent differences in dynamic structures between humans and humanoid robots, directly abstracting human motions into robot actions is often infeasible \cite{6651585}. Much recent work has employed Inverse Kinematics (IK) \cite{darvish2019whole, baek2024human} or neural network-based optimization \cite{choi2021self, yan2023imitationnet, yagi2024unsupervised}. However, these methods either focus on teleoperation of decomposed robot parts (e.g., upper-body teleoperation) or rely on configuration-specific manual adjustments and parameterization, entailing significant engineering overhead and lacking universality. Furthermore, while optimization-based IK solutions offer a degree of automation, the multi-solution problem arising from redundant Degrees of Freedom (DoF) can lead to kinematically infeasible or discontinuous sequences between consecutive frames \cite{darvish2023teleoperation}. These approaches also struggle with efficient, real-time conversion of Motion Capture (MoCap) data. Consequently, there is an urgent need for an efficient, general-purpose, and executable data processing pipeline to transform diverse human motion datasets (e.g., AMASS \cite{AMASS:ICCV:2019}, MotionX \cite{lin2023motion}) into physically feasible robot demonstration datasets that preserve the essence of the motion and are suitable for various humanoid configurations and DoFs.

Even with high-quality robot demonstration data, training a policy that can accurately imitate demonstrated actions, exhibit robustness to perturbations, and generalize to new reference motions remains a significant challenge. Humanoid robots are inherently unstable systems, and direct Behavior Cloning often suffers from covariate shift and lacks effective incorporation of mechanical constraints \cite{saeedvand2019comprehensive}. Although DAgger \cite{ross2011reduction} and Generative Adversarial Imitation Learning (GAIL) \cite{ho2016generative} have proposed effective improvements for such issues, they can still be constrained by the quality of the demonstration data. Conversely, algorithms relying on traditional RL face difficulties in reward function design, vast exploration spaces, and often unclear convergence paths. Recent studies attempting to combine RL and IL to enhance performance \cite{peng2021amp, he2024learning, ze2025twist} still encounter limitations such as a focus on single robot configurations, stringent MoCap requirements, and difficulties in reproducing results across similar tasks. Moreover, as RL backbones are often overly simplistic Multi-Layer Perceptrons (MLPs), they struggle to effectively capture the mapping between reference states and current agent states, especially when there is a large initial discrepancy between these distributions. Therefore, designing a comprehensive RL+IL learning environment that can effectively balance demonstration data with actual observations is also key to success.

To systematically address the aforementioned challenges, we propose the GBC (Generalized Behavior Cloning) framework, \textbf{a very first, comprehensive universal behavior cloning framework designed for whole-body imitation across humanoid robots with heterogeneous morphologies and varying DoFs}. Our work makes the following primary contributions:
\begin{enumerate}
    \item \textbf{A real-time, differentiable MoCap data processing pipeline:} This pipeline, through our optimized differentiable IK network and a series of carefully designed post-processing and data augmentation procedures, converts MoCap data—either offline or online—into physically feasible, high-quality demonstration datasets that are adaptive to different robot configurations.
    \item \textbf{A powerful RL+IL algorithm with an improved backbone network:} We propose the DAgger-MMPPO algorithm as the core of GBC. It achieves efficient action imitation through two-stage RL and supports the integration of other effective IL methods such as Adversarial Motion Priors (AMP). Based on our designed MMTransformer network, this algorithm treats the robot's egocentric observations and reference motion states as distinct modalities, thereby modeling the relationship between them. The resulting policy can not only imitate human actions when provided with demonstration data but also retains the capability to follow velocity commands in a human-like manner when such guidance is absent, while also being applicable to other PPO-based RL tasks.
    \item \textbf{An easy-to-deploy, computationally efficient training platform:} The GBC framework is developed as an extension of Isaac Lab and Isaac Sim, enabling efficient GPU-accelerated scheduling and training support for all implemented algorithms. It also integrates additional enhancements such as curriculum learning \cite{bengio2009curriculum}, physics-based assistance, and augmented randomization strategies to optimize imitation training outcomes. Users only need to prepare configuration files specific to their robot model to execute an efficient workflow from dataset generation to algorithm training. Our framework is open-sourced on GitHub with detailed documentation support and deployment guide\cite{GBC_Framework_2025}.
\end{enumerate}

To validate the generality of our framework, we tested the data processing pipeline's versatility and the imitation learning algorithm's effectiveness on Unitree G1, Unitree H1-2, Fourier GR1, and Turin humanoid configurations. We also compared the performance of our algorithm against traditional methods in both basic RL and IL tasks. Experiments demonstrate that our GBC framework can learn general whole-body imitation policies for various robot configurations, exhibiting good adaptation and transfer capabilities to novel actions. Furthermore, sim-to-sim experiments confirm that the trained policies are deployable in realistic physics simulations, indicating strong application potential.

\section{Related Works}
\subsection*{Motion Retargeting and Teleoperation} 
Motion retargeting, which refers to the conversion of human-derived motion data into executable motion commands for humanoid robots, is fundamental for addressing the physical infeasibility of directly applying human actions to robots and forms a cornerstone for humanoid teleoperation \cite{darvish2023teleoperation}. Methods for retargeting in recent years encompass several approaches, including those based on Inverse Kinematics (IK), optimization, and machine learning. IK-based methods \cite{ayusawa2015motion,zhang2018real,penco2018robust,aberman2020unpaired, darvish2019whole} are relatively straightforward but suffer from multi-solution ambiguities and often exhibit suboptimal real-time performance. Optimization techniques developed to address the deficiencies of basic IK, such as those in \cite{6651585,bin2015kinodynamically,di2016multi,liang2021dynamic}, can achieve higher precision but typically require configuration-specific adjustments and parameterization. More recently, deep learning approaches, whether data-driven or reinforcement learning-based \cite{choi2020nonparametric,zhang2021human,yan2023imitationnet,yagi2024unsupervised}, have demonstrated significant potential in terms of generalization and real-time capability, offering lower-cost and higher-efficiency solutions for real-time teleoperation. Compared to teleoperation schemes reliant on specialized Motion Capture (MoCap) suits \cite{ben2025homie, ze2025twist}, those employing RGB cameras with VR assistance \cite{he2024learning,he2024omnih2o}, or RGB cameras alone \cite{fu2024humanplus}, our work proposes a more unified interface. For any MoCap format compliant with the AMASS\cite{AMASS:ICCV:2019} data specification (regardless of the capture modality), our data processing pipeline provides a real-time retargeting interface based on a learning-driven differentiable IK network, complemented by optimization-based post-processing. This approach aims to mitigate inaccuracies and dynamic discrepancies inherent in raw MoCap data. Furthermore, our pipeline supports the conversion of human MoCap datasets adhering to this specification into robot training datasets, thereby enabling not only imitation learning but also paving the way for subsequent motion generation training based on datasets like HumanML3D \cite{Guo_2022_CVPR}.

\subsection*{Reinforcement and Imitation Learning for Humanoids}
Given the complexities of modeling traditional controllers for human imitation tasks, researchers predominantly employ Reinforcement Learning (RL) or Imitation Learning (IL) to achieve human-like control of humanoid robots. RL has demonstrated exceptional performance in tasks such as robot walking, jumping, running, and terrain adaptation \cite{chen2024learning, radosavovic2024real, zhang2024wococo, cheng2024expressive}, but its focus is often on solving specific tasks rather than mimicking nuanced human behaviors. Behavior Cloning (BC) algorithms like DAgger \cite{ross2011reduction} and Generative Adversarial Imitation Learning (GAIL) \cite{ho2016generative} enable policy derivation from demonstration data, facilitating data-driven strategy learning and control. Recent investigations into Behavior Transformers (BeT) \cite{shafiullah2022behavior, lee2024behavior} have highlighted the potent capabilities of Transformer networks \cite{vaswani2017attention} as policy backbones for modeling complex relationships between observations and reference motions. However, pure IL methods remain constrained by the scope of the demonstration data, exhibiting out of distribution (O.O.D.) issues while migrating to other tasks and suboptimal robustness against perturbations. Consequently, strategies combining RL with IL \cite{2018-TOG-deepMimic,peng2021amp,he2024learning} are generally more effective in enabling robots to master complex human motion patterns. Our work extends existing RL+IL paradigms by synergizing IL with an on-policy method, Proximal Policy Optimization (PPO) \cite{schulman2017proximal}. We employ a specially designed Transformer backbone to replace commonly used MLP-based policies, enabling the trained policy to more effectively learn the correlations between demonstration data and environmental observations. This approach supports both RL control (without reference) and reference-based motion imitation, and can be organically integrated with existing IL optimization techniques to achieve more efficient and precise humanoid policy training.

\subsection*{On-Policy Training Optimization Techniques}
In practical on-policy training, directly imitating target motions can be challenging due to the large exploration space agents must navigate. Furthermore, the sim-to-real gap often prevents policies trained in simulation from being directly deployed on physical hardware. To address these issues, various training optimization techniques are commonly introduced. To mitigate discrepancies between simulation and reality, Domain Randomization (DR)\cite{tobin2017domain} is frequently employed, involving the randomization of environmental parameters, robot kinematic and dynamic properties, and the introduction of external disturbances to enhance policy generalization \cite{ peng2018sim,bousmalis2018using, vuong2019pick}. To counteract the problem of sparse rewards in initial training phases, Curriculum Learning (CL) \cite{bengio2009curriculum, jiang2015self, florensa2017reverse} is introduced, which progressively increases the learning difficulty to first guide the policy towards a convergent trajectory and then to refine its performance. Additional optimizations, such as incorporating symmetry or energy constraints \cite{yu2018learning, abdolhosseini2019learning}, and randomizing sampling locations or training data to prevent policy degradation \cite{2018-TOG-deepMimic, peng2021amp}, contribute to learning more stable policies and avoiding infeasible control outputs. Building upon these established methodologies, our framework introduces further refinements. We incorporate a specific form of CL for motion tracking targets and a unique physics-based assistance curriculum. Within our PPO framework, we implement randomization of reference action types and initial starting states, alongside integrated symmetry and energy-inspired constraints. These design allow improvements in both training efficiency and the ability to transfer learned skills across different environmental conditions.

\section{Core Modeling and Algorithmic Foundations}\label{sec:algorithm}

To make our framework applicable to humanoid robots with heterogeneous morphologies, our dataset preparation pipeline is organized into three stages:
\emph{First}, We fit the human mesh description to the target robot by solving a
shape–calibration problem.  
\emph{Second}, we use a differentiable inverse-kinematics (IK) network to map motion-capture
(MoCap) poses to robot joint angles under kinematic constraints.  
\emph{Third}, we use a post-processor to smooth the joint stream, augment it with
reference signals, and apply certain data argumentation techniques. 

On top of the generated expert dataset we design a demonstration-oriented reinforcement learning backbone and introduce several modifications to the standard Proximal Policy Optimization (PPO) algorithm \cite{schulman2017proximal}, alleviating the suboptimal convergence behavior observed with conventional MLP or RNN based PPO policies. The improved framework can be seamlessly combined with established imitation-learning techniques, such as DAgger \cite{ross2011reduction} or Adversarial Motion Priors (AMP) \cite{peng2021amp}, to further boost training efficiency. We now introduce the algorithm implementation in detail.

\subsection{Differentiable IK and Dataset Preparation Pipeline}\label{subsec:pipeline}

\begin{figure}[htbp]
    \centering
    \includegraphics[width=0.95\linewidth]{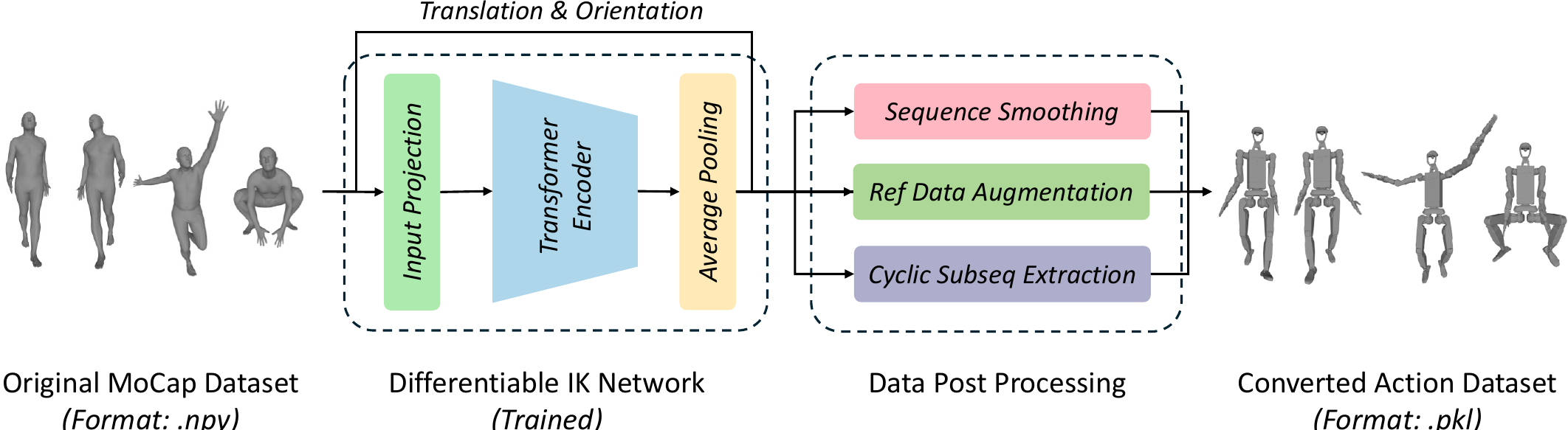}
    \caption{GBC data processing pipeline. MoCap data (angle-axis representation) are converted by differentiable IK network, filtered and augmented along with global translation and orientation data, and stored as humanoid expert dataset.}
    \label{fig:data_pipeline}
\end{figure}

To enable humanoid robots to imitate whole body human motion, we first require low latency demonstration data.
Such data are usually obtained by retargeting optical MoCap
trajectories onto the robot kinematic chain.
For example, the \emph{Unitree Retargeting Dataset}\cite{lv_2025_lafan1_retargeting} derives its joint commands
from the LaFAN1 dataset \cite{harvey2020robust} by combining an
\emph{Interaction Mesh} representation \cite{6651585} with conventional
inverse kinematics (IK). %
Other pipelines like \emph{Human2Humanoid} \cite{he2024learning} take a more direct route:
it formulates IK as an optimization over end--effector positions and solves it by gradient descent to obtain the
best‐fitting joint angles. However, both classes of methods face practical limitations.
Detailed IK has to be re-parameterized for a new robot
morphology, leading to substantial engineering overhead.
In contrast, per-frame optimization over end effector constraints is easy to
implement, but its solution is not guaranteed to satisfy the full kinematic
feasibility or maintain continuity between frames, and often fails to capture intrinsic mapping regularities from
MoCap space to robot joint space.
As a result, neither approach achieves the
\emph{real-time, high-throughput MoCap‐to‐action conversion}
required for closed-loop imitation learning. 


Although IK on highly–redundant humanoid robots is in
general a \emph{multi–solution} problem, we argue that, for the purpose of
\emph{retargeting a fixed, continuous trajectory}, the mapping from a human
pose to the corresponding robot configuration can be treated as injective.
Let the human–motion space defined in angle--axis form be
\(\mathbb{P}\subset\mathbb{R}^{J_{\mathrm{hm}}\times 3}\) and the humanoid
action space be \(\mathbb{A}\subset\mathbb{R}^{J_{\mathrm{hn}}}\).
For any human pose \(p\in\mathbb{P}\) there exists a
\emph{unique} robot joint vector \(a\in\mathbb{A}\), yielding a bijective\footnote{%
Strictly speaking the map is injective on the support of the demonstration
data; surjectivity is not required.}
function
\(f:\mathbb{P}\rightarrow\mathbb{A}\)
that can be learned by a neural network. Thus we can introduce a differentiable \emph{light-weight neural regressor} to implement a simple and versatile system with real-time retargeting capabilities. Here we demonstrate how this model is created and how the data processing pipeline works.

\subsubsection{Parametric Model Fitting}
The AMASS dataset~\cite{AMASS:ICCV:2019} stores motion trajectories using the
\textsc{SMPL+H} body model\cite{SMPL:2015, MANO:SIGGRAPHASIA:2017}, which comprises \(22\) body joints and \(30\) hand
joints.  Rotations are represented in angle--axis form and the database
contains more than \(11{,}000\) annotated sequences.
A 16-dimensional PCA latent vector
\(\boldsymbol{\beta}\in\mathbb{R}^{16}\) encodes \emph{static morphology}
(height, limb length, body shape), while \(\boldsymbol{\delta}\)
(\textit{dmpls}) captures dynamic soft-tissue deformations.
Together with joint angles \(p\), these parameters are fed into the
differentiable forward-kinematics routine provided by AMASS,
\(\operatorname{FK}_{\mathrm{hm}}(p,\boldsymbol{\beta},
\boldsymbol{\delta})\). Also, since the robots have different heights, we use a certain scale factor $\boldsymbol{\alpha}$ to resize SMPL+H skeletons to humanoid's size.

To interface the mocap data with a target humanoid, we calibrate the shape
parameters so that human and robot end-effectors coincide for a small set of
paired reference poses
\(\mathcal{M}\!=\!\{(p^{m},a^{m})\}\),
\(p^{m}\!\in\!\mathbb{P}\),
\(a^{m}\!\in\!\mathbb{A}\)
(e.g.\ neutral stance, \textsc{T}-pose, user–specified key frames).
Formally, we solve
\begin{equation}
\label{eq:shape_opt}
\underset{\boldsymbol{\alpha},\,\boldsymbol{\beta},\,\boldsymbol{\delta}}\arg\min
\sum_{m\in\mathcal{M}}
\bigl\|
\boldsymbol{\alpha}\operatorname{FK}_{\mathrm{hm}}
  \!\bigl(p^{m},\boldsymbol{\beta},\boldsymbol{\delta}\bigr)
-
\operatorname{FK}_{\mathrm{hn}}
  \!\bigl(a^{m}\bigr)
\bigr\|_{2},
\end{equation}
where \(\operatorname{FK}_{\mathrm{hn}}(\cdot)\) denotes the robot’s forward
kinematics.
The larger \(\lvert\mathcal{M}\rvert\), the more accurately the fitted
morphology matches the robot.



\begin{figure}[htbp]
    \centering
    \includegraphics[width=0.65\linewidth]{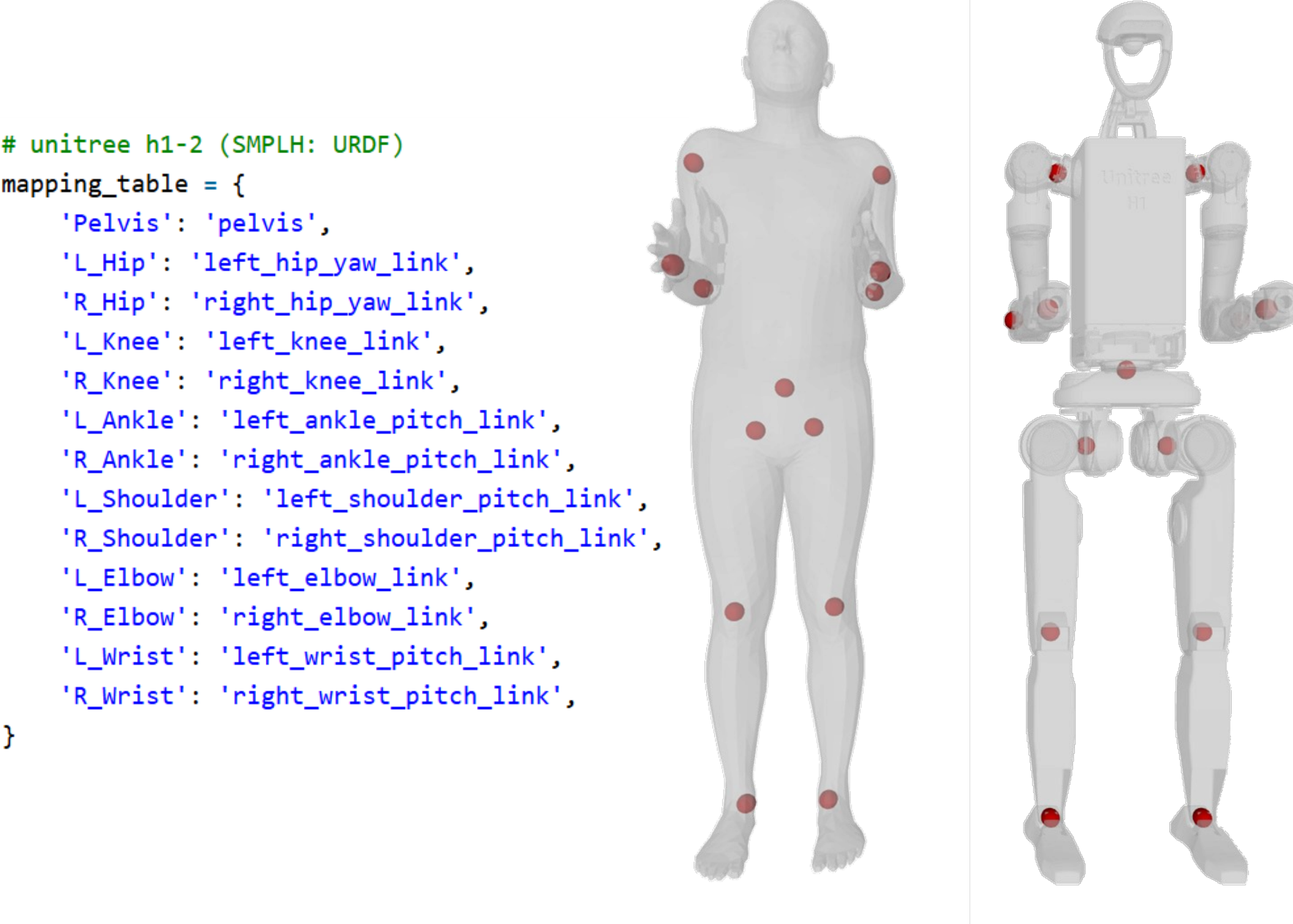}
    \caption{Joint mapping from human to humanoid defined for Unitree H1-2.}
    \label{fig:joint_mapping_table}
\end{figure}

\begin{figure}[htbp]
  \centering
  \subfigure[Shape Fit for H1-2.]{%
    \includegraphics[width=0.48\linewidth]{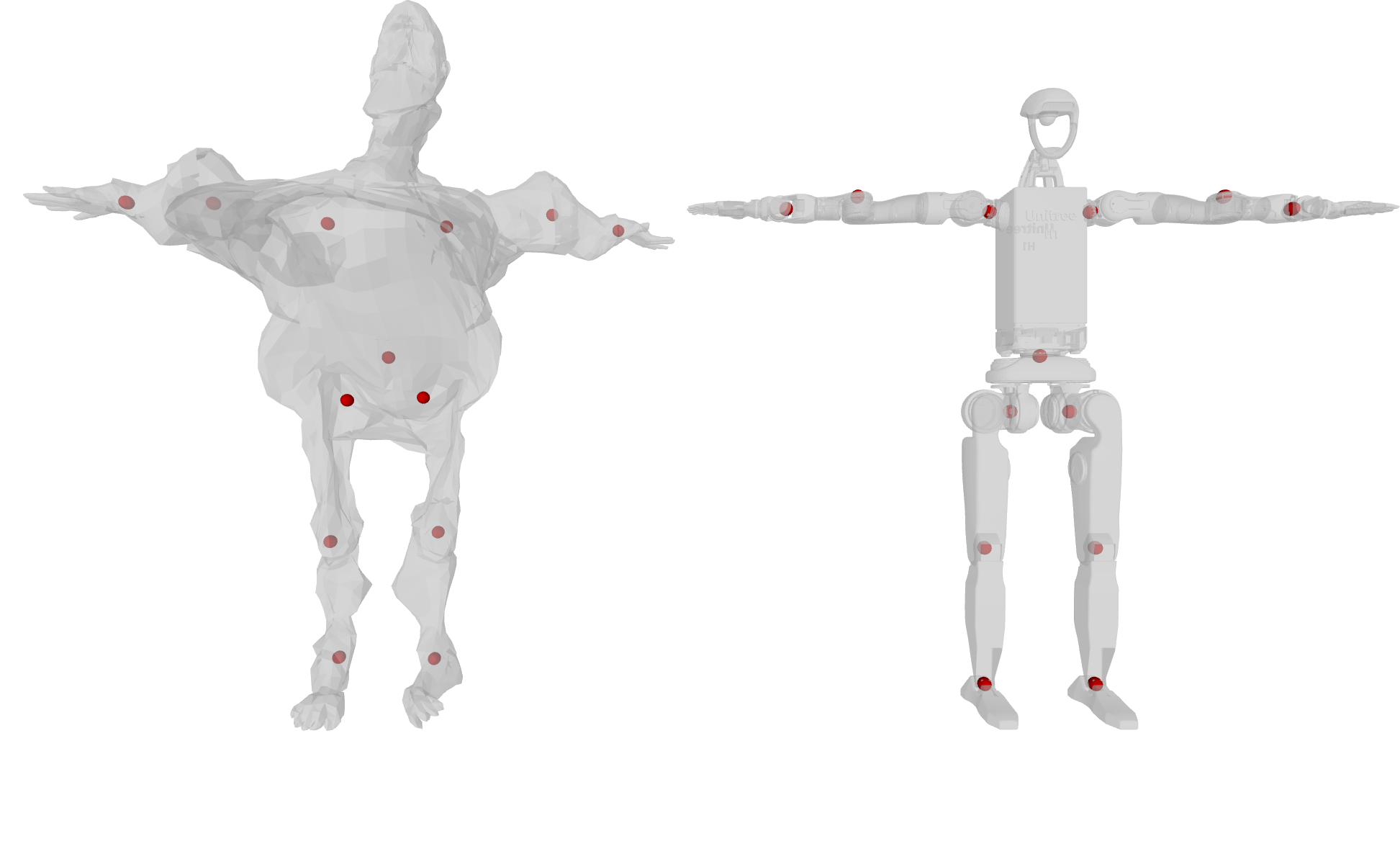}}
  \hfill
  \subfigure[Shape Fit for Turin HN.]{%
    \includegraphics[width=0.48\linewidth]{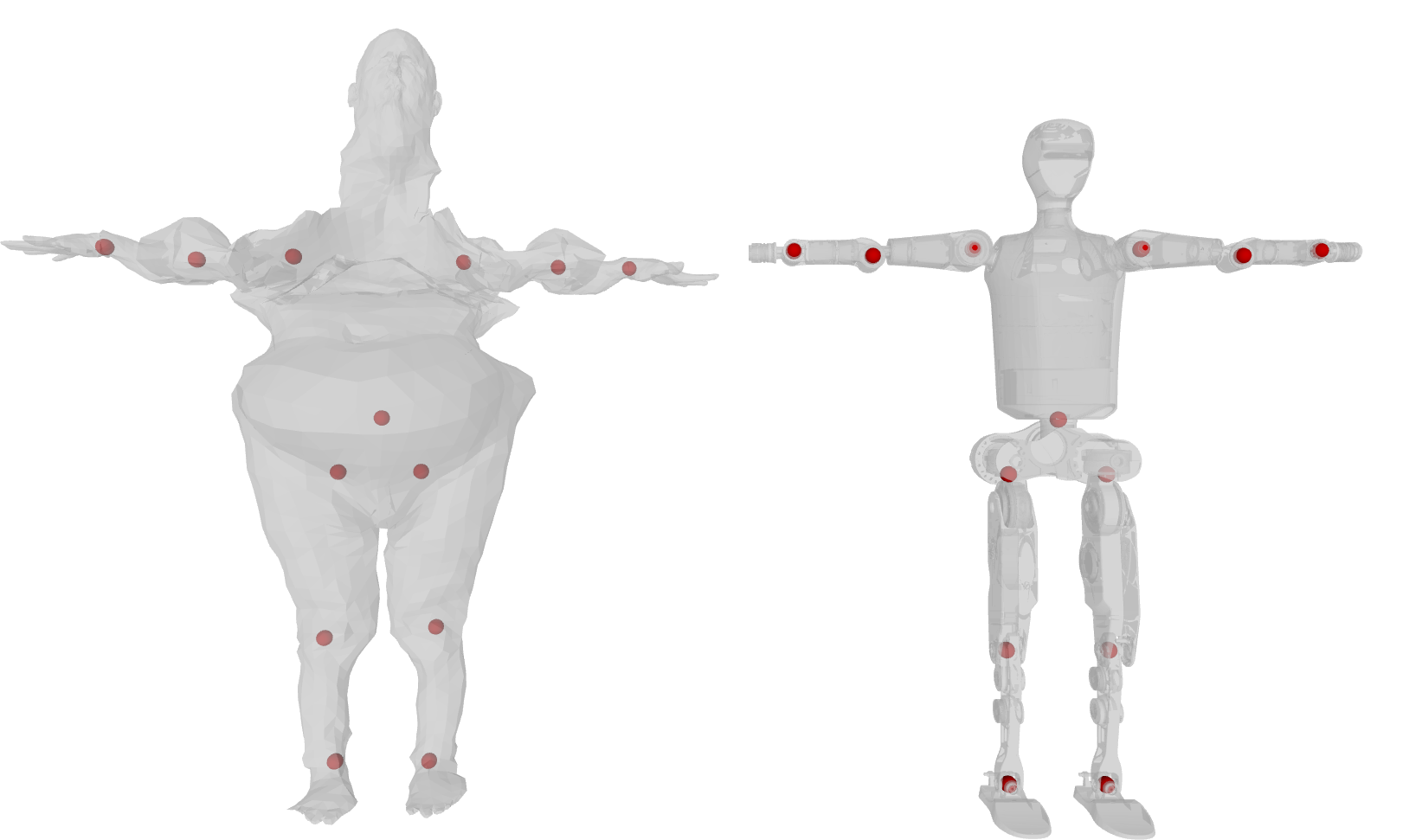}}

  \caption{SMPL+H shape fit for different humanoid architectures (left: fitted SMPL+H, right: URDF).}
  \label{fig:SMPL+Hfitter}
\end{figure}

Equation~\eqref{eq:shape_opt} is optimized in PyTorch; the URDF parser
automatically constructs the robot kinematic tree and establishes a
one-to-one mapping between human and robot end-effectors with user-provided joints name mapping table like Figure~\ref{fig:joint_mapping_table}.
Once \(\boldsymbol{\beta}^{\star},\boldsymbol{\delta}^{\star}\) are found,
\(\operatorname{FK}_{\mathrm{hm}}(\cdot,\boldsymbol{\beta}^{\star},
\boldsymbol{\delta}^{\star})\)
becomes a drop-in differentiable surrogate for the robot morphology,
closing the chain
\(\mathbb{P}\;\xrightarrow{f}\;\mathbb{A}\).
Unlike hand-engineered Euler-angle retargeting, the proposed shape fitting
requires \emph{no robot-specific modelling}, enabling strong morphological
generalization.
Figure~\ref{fig:SMPL+Hfitter} visualizes the calibrated meshes for two
distinctive humanoid platforms.

\subsubsection{IK Modeling and Regression}

We employ a light–weight \emph{Transformer encoder} to approximate~\(f\), where its architecture is inspired by Vision Transformers\cite{dosovitskiy2020image} to embed joint features to latent space.
To keep the entire pipeline differentiable, we implement a
robot--specific forward-kinematics (FK) library
\(\operatorname{FK}_{\mathrm{hn}}\!\left(\cdot\right)\) that supports pytorch backward propagation
and reuse the human FK provided by the AMASS prior
\(\operatorname{FK}_{\mathrm{hm}}\!\left(\cdot\right)\).
The model is trained on frame–wise samples
from the AMASS dataset; the primary objective minimizes the Euclidean
distance between the robot’s end-effector locations and those of the source
mocap frame:

\begin{equation}\label{eq:dist}
\mathcal{L}_{\text{dist}}(p)=
\bigl\|
\operatorname{FK}_{\mathrm{hn}}\!\bigl[f(p)\bigr]\;-\;
\operatorname{FK}_{\mathrm{hm}}(p)
\bigr\|_{2}.
\end{equation}

Equation~\eqref{eq:dist} by itself reduces the problem to a
differentiable IK optimization and, therefore, inherits the usual pathologies
(output discontinuities, dynamically infeasible poses).  To address this, we thus introduce
three auxiliary losses.


\paragraph*{(a) Joint-limit loss}
Let \(a_{\min},a_{\max}\in\mathbb{R}^{J_{\mathrm{hn}}}\) be the lower and
upper joint limits parsed from the URDF.
Violations are penalised by
\begin{equation}\label{eq:limit}
\mathcal{L}_{\text{limit}}(p)=
\bigl\|
\max\!\bigl(a_{\min}-f(p),\,0\bigr)\;+\;
\max\!\bigl(f(p)-a_{\max},\,0\bigr)
\bigr\|_{2},
\end{equation}
where the \(\max\) is applied element-wise.%
\footnote{If the URDF limits are unrealistically wide, we additionally bound
each joint by empirical human ranges, e.g.\ \(\pm180^{\circ}\) for shoulders.}

\paragraph*{(b) Action–disturbance loss}
Assume the pose spaces are bounded by
\(\sup\mathbb{P}=\sup\mathbb{A}=2\pi\) and
\(\inf\mathbb{P}=\inf\mathbb{A}=0\).
Because the Transformer is differentiable and has bounded gradients at
convergence, an ideal mapping \(f\) should be \emph{Lipschitz continuous}.
We enforce a small Lipschitz constant by perturbing the input with zero–mean
Gaussian noise \(\delta p\sim\mathcal{N}\!\bigl(0,\sigma(2\pi)^{2}\bigr)\),
\(\sigma\ll 1\), and minimizing
\begin{equation}\label{eq:disturb}
\mathcal{L}_{\text{disturb}}(p)=
\frac{\bigl\|
\operatorname{FK}_{\mathrm{hn}}\![f(p)]-
\operatorname{FK}_{\mathrm{hn}}\![f(p+\delta p)]
\bigr\|_{2}}
{\|\delta p\|_{2}} .
\end{equation}

This implementation along with spectral normalization assist network to minimize the impact of  both high and low frequency disturbance to network input.
\paragraph*{(c) Symmetry loss}
Let \(\mathcal{S}\) denote a reflection operator about the humanoid’s sagittal
plane.  For a human pose \(p\) we write
\(p'=\mathcal{S}(p)\) and analogously apply \(\mathcal{S}\) to robot
end-effector positions.
Ideally, retargeting should preserve bilateral symmetry, leading to
\begin{equation}\label{eq:sym}
\mathcal{L}_{\text{sym}}(p)=
\bigl\|
\mathcal{S}\!\bigl(\operatorname{FK}_{\mathrm{hn}}\![f(p)]\bigr)\;-\;
\operatorname{FK}_{\mathrm{hn}}\!\bigl[f(p')\bigr]
\bigr\|_{2}.
\end{equation}

\paragraph*{Composite objective}
The total loss combines the four terms with empirically tuned weights
\(\lambda_{\text{dist}}\!>\!\lambda_{\text{limit}}\!>\!
\lambda_{\text{disturb}}\approx\lambda_{\text{sym}}\):
\[
\mathcal{L}_{\text{total}}=
\lambda_{\text{dist}}\mathcal{L}_{\text{dist}}+
\lambda_{\text{limit}}\mathcal{L}_{\text{limit}}+
\lambda_{\text{disturb}}\mathcal{L}_{\text{disturb}}+
\lambda_{\text{sym}}\mathcal{L}_{\text{sym}}.
\]

Additionally, consider that some joints like knees possess a single valid degree-of-freedom (DoF), we allow users to specify these joints and add the difference between the joint angle component and the network output for that DoF as a loss function. 

\paragraph*{Training protocol}
Frames are sampled from AMASS with 15 to 30 frame intervals (1-2 fps) and split \(80{:}20\) into
training/validation sets.
The Transformer encoder takes
\(\smash{J_{\mathrm{hm}}\times3}\) angle–axis vectors\footnote{For body retargeting only, $\smash{J_{\mathrm{hm}}}=22$; for retargeting with hands, $\smash{J_{\mathrm{hm}}}=22+30=52$.}, projects them to a
\(\textit{seq\_len}\!\times\!d_{\text{model}}\) token sequence, and finally
projects to the robot joint dimension followed by mean pooling.
A two–layer model with \(d_{\text{model}}=64\) contains fewer than
0.1 M~parameters and runs in real time even on a CPU.

\subsubsection{Sequence Retargeting and Post-Processing}
The differentiable IK described above maps the original angle–axis stream
onto the robot’s joint space, yet the raw output cannot be executed by the
humanoid without additional processing.  In particular

\begin{itemize}
\item sparse temporal sampling and the limited capacity of the light-weight
      network introduce high–frequency jitter;
\item locomotion sequences that keep a human subject balanced are not
      necessarily dynamically feasible for a robot \cite{6651585}; and
\item AMASS clips vary in length, which impedes stable training in
      fixed-horizon RL environments.
\end{itemize}

We therefore apply a three-stage post-processing routine: \emph{temporal
smoothing}, \emph{reference-signal augmentation}, and
\emph{cyclic-subsequence extraction}.

\begin{figure}[htbp]
    \centering
    \includegraphics[width=0.95\linewidth]{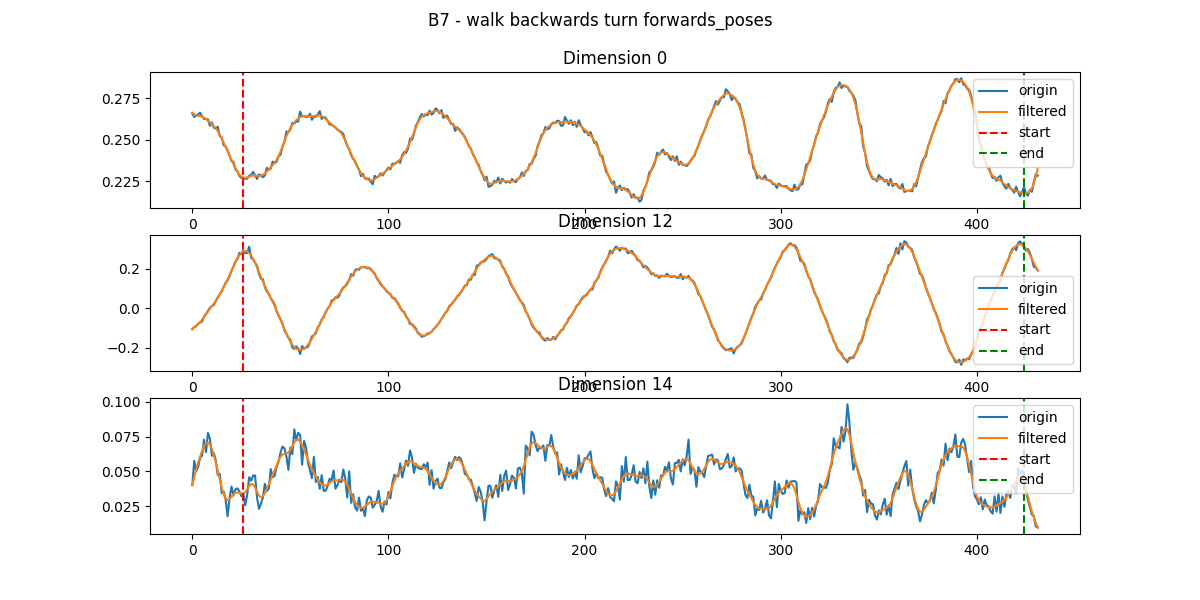}
    \caption{Example of filtering and cyclic sequence extraction, this motion is from AMASS/ACCAD.}
    \label{fig:filter_and_cyclic_subseq}
\end{figure}

\paragraph*{Temporal smoothing}
For off-line demonstrations the user specifies a target frame rate
\(f_{\text{tgt}}\); the retargeted joint sequence is re-sampled to that rate
via SLERP interpolation, followed by a Butterworth low-pass filter whose
cut-off frequency is set to \(0.1\,f_{\text{tgt}}\).
For real-time streaming we employ a causal \(2{:}6{:}2\) moving-average
filter.
Figure~\ref{fig:filter_and_cyclic_subseq} illustrates the effect on a
randomly chosen joint channel.

\paragraph*{Reference-signal augmentation}
To furnish the RL agent with richer supervision we compute additional
ground-truth signals---\emph{only for the critic} to keep the policy compact.
Given global translation \(\mathbf{t}_t\) and orientation
\(\mathbf{R}_t\) from AMASS, the robot-frame linear velocity
\(\dot{\mathbf{t}}^{\text{rob}}_t\), angular velocity
\(\boldsymbol{\omega}^{\text{rob}}_t\), IMU gravity vector and joint
velocities are obtained through
\(\operatorname{FK}_{\mathrm{hn}}\).
Foot–ground contact is detected when the foot end-effector has
\(\|\dot{\mathbf{x}}_{\text{foot}}\|\!<\!0.05\mathrm{\,m/s}\) and its height
is below a certain threshold; the resulting binary phase is encoded as a
cosine–sine pair to match common locomotion controllers.  Because the robot
is randomized in position and yaw at episode start, we store only
\emph{differential} quantities, which are recomputed online from the
reset state.  A complete list of reference channels is given in Appendix~A.

\paragraph*{Cyclic-subsequence extraction}
Many skills in AMASS---walking, running, jumping---are approximately periodic.
To unify episode lengths we search each retargeted sequence for the longest
cycle.  Let \(\mathbf{q}_1,\dots,\mathbf{q}_n\) be the joint sequence and
let \(\varepsilon\) be a user-defined tolerance.
A KD-tree is built on the joint vectors; we locate the first pair
\((i,j)\) with
\(\lvert j-i\rvert \ge 0.2\,n\) and
\(\|\mathbf{q}_i-\mathbf{q}_j\|_2\le\varepsilon\),
employing start/end–distance pruning and early termination.
The average complexity is \(O(\log n)\).
The cyclic index pair \((i, j)\) (red/green dashed lines in
Fig.~\ref{fig:filter_and_cyclic_subseq}).
Processed demonstrations are stored in \texttt{pickle} format for
high-throughput loading during RL; usage details are provided in
Section~\ref{sec:system} and the Supplementary material.

\subsection{Motion Mimic Transformer (MMTransformer) Backbone}

\begin{figure*}[htbp]     
    \centering
    \includegraphics[width=0.90\textwidth]{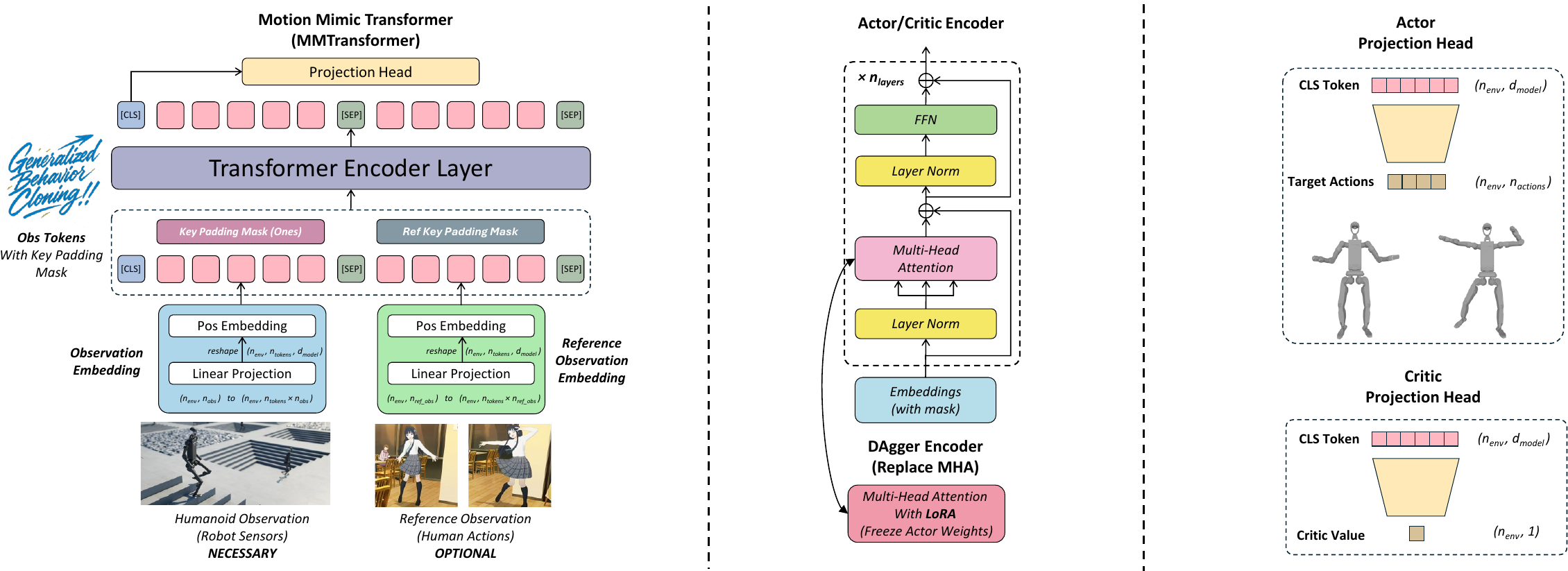}  
    \caption{MMTransformer architecture. MMTransformer follows a BERT style encoder design, and treats humanoid observations and reference observations (expert data) as different modalities. Reference observations are masked out in environment dimensions where reference is no longer available. The transformer encoder layer freezes its main weights with LoRA during DAgger step teacher-student distillation. The projection head for actor and critic backbone project out CLS token to either target actions or critic value.} 
    \label{fig:mmtransformer}
\end{figure*}

\begin{figure}[htbp]     
    \centering
    \includegraphics[width=0.98\linewidth]{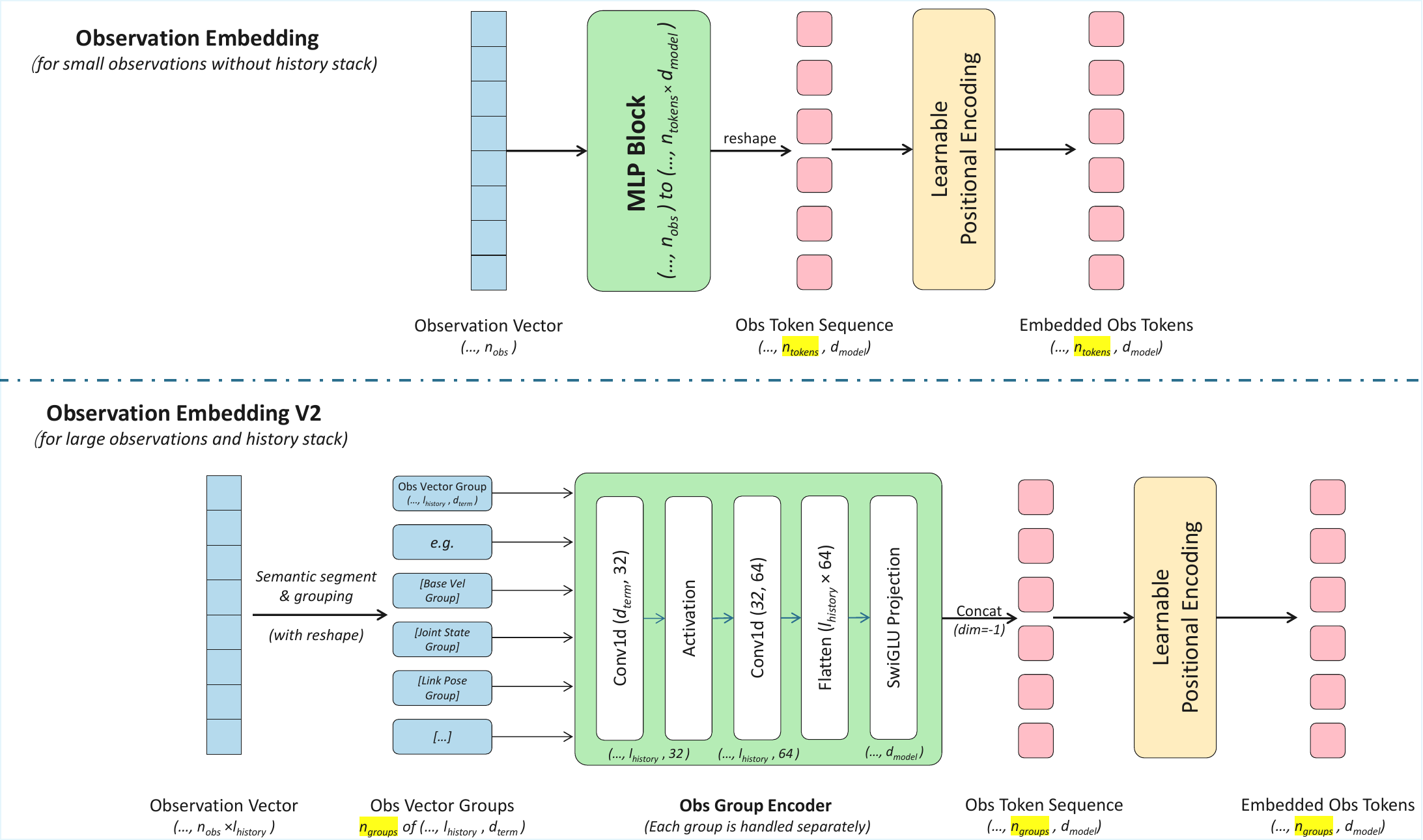}  
    \caption{Embedding layers for MMTransformer. For simple observations with lesser dimensions and no history stacks, simple projection to reshape embedding (\textbf{Observation Embedding}) structure is useful enough. But for multiple observations and reference observations with higher dimensions and history stacking, this function is not enough. We introduce \textbf{Observation Embedding V2} to group observations by their categories (e.g. We tend to group linear velocity, angular velocity together as ``base state group'', joint positionss together as ``joint state group'', etc.), handle history input with convolutional layer, model the in-group relation with SwiGLU projection used in Llama\cite{touvron2023llama} and let transformer model the between-group relations.} 
    \label{fig:obsembedding}
\end{figure}

After the expert dataset has been prepared, the next objective is to train a
RL agent that can track the reference motion and
internalize transferable motor skills.  
The default PPO implementation employs a
\emph{multi-layer perceptron}~(MLP) policy, which is computationally
efficient but proved inadequate for high-dimensional references: our pilot
experiments showed poor correlation between the observation state vector
\(o_t\) and the reference input \(\hat{o}_t\).
Replacing the MLP with a \emph{recurrent neural network} (RNN) improves
expressive power~\cite{ni2021recurrent}, yet the recurrence mainly models
temporal continuity and still struggles to capture the structured
relationship between \(o_t\) and \(\hat{o}_t\).
Furthermore, in \textsc{Isaac Lab}\footnote{Up to now (2025.6), newest version 2.1 still concatenates timed observations by terms.} the observation interface does
not expose the stacked history directly as required by sequence-to-sequence encoders,
making Decision-Transformer–style methods~\cite{chen2021decision} less
practical.

To overcome these limitations we adopt a \emph{Transformer-based policy}
that treats the observation and the reference as separate token streams and
aligns them through attention, since previous researches have revealed its great potential of multi-modality processing\cite{carion2020end, kim2021vilt}. We refer to this backbone as the
\textbf{Motion-Mimic Transformer (MMTransformer)}; its architecture and
calculation details are presented as follows.

The \textbf{MMTransformer} adopts an \emph{Encoder-Only} architecture based on the BERT configuration \cite{devlin2019bert}. It processes the observation \(o_t\) and the reference input \(\hat{o}_t\) (termed \emph{reference observation} to signify its connection to the observation) as \emph{distinct modalities}. We incorporate a \emph{binary mask} \(b_{env} \in \{0, 1\}\) for each environment's \(\hat{o}_t\), where \(b_{env}=1\) indicates its availability and \(b_{env}=0\) indicates non-availability. Both \(o_t\) and \(\hat{s}_t\) are transformed into \emph{Observation Tokens} via separate \emph{Embedding layers} demonstrated in Fig. \ref{fig:obsembedding}. These Embedding layers are specifically designed to meet the \emph{sequential input requirements} of Transformers.
When the number of observation types is small (insufficient to form a robust sequence), an observation \(x \in \mathbb{R}^{\text{dim}_{\text{in}}}\) (representing either \(o_t\) with \(\text{dim}_{\text{in}} = \text{obs\_dim}\) or \(\hat{o}_t\) with \(\text{dim}_{\text{in}} = \text{ref\_obs\_dim}\)) is projected and reshaped. Specifically, its token embedding \(E_x \in \mathbb{R}^{\text{seq\_len} \times d_{\text{model}}}\) is obtained via \(E_x = \text{Reshape}(\boldsymbol{W}x + \boldsymbol{b})\), where the linear transformation with weight \(\boldsymbol{W} \in \mathbb{R}^{(\text{seq\_len} \cdot d_{\text{model}}) \times \text{dim}_{\text{in}}}\) and bias \(\boldsymbol{b} \in \mathbb{R}^{\text{seq\_len} \cdot d_{\text{model}}}\) projects \(x\) to a vector of dimension \(\text{seq\_len} \cdot d_{\text{model}}\).
Conversely, when there are numerous observation types, we set \(\text{seq\_len}\) to be the number of these types. Terms such as linear velocity, angular velocity, joint position, and joint velocity are individually embedded by projecting each to \(d_{\text{model}}\), and these resulting vectors are subsequently \emph{concatenated} (stacked) to form the final \((\text{seq\_len}, d_{\text{model}})\) tensor. The Embedding layer for the reference observation \(\hat{s}_t\) follows an \emph{identical structure but operates independently}. If the reference input \(\hat{s}_t\) is `None' during operation, its corresponding Embedding layer is not registered, thereby reducing memory consumption. To incorporate information about the token order within the sequence, we utilize \emph{learned positional embeddings} with shape \((\text{seq\_len}, d_{\text{model}})\). These positional embeddings are then added to the corresponding token feature embeddings before being fed into the Transformer encoder.

The transformed observation tokens from \(o_t\) (denoted \(E_{o_t}\)) and \(\hat{o}_t\) (denoted \(E_{\hat{s}_t}\)) are separated by a \texttt{[sep]} token. A \texttt{[cls]} token is prepended to the beginning of the overall sequence (e.g., \texttt{[cls]} \(E_{o_t}\) \texttt{[sep]} \(E_{\hat{o}_t}\)), enabling the model to learn the relationships between the observation and the reference observation through this \texttt{[cls]} token.
For a reference input \(\hat{o}_t\) (per environment), which is embedded to a sequence of length \(\text{ref\_seq\_len}\), a corresponding \emph{key\_padding\_mask} is generated to \emph{block attention computations} for non-available reference tokens. This mask is produced by taking a sequence of all ones (of length \(\text{ref\_seq\_len}\)) and performing an element-wise multiplication with a per-environment scalar mask \(m'_j = (1-b_{env,j})\). Thus, if the reference is available for environment \(j\) (\(b_{env,j}=1\)), \(m'_j=0\), resulting in a zero padding mask (attend to tokens). If unavailable (\(b_{env,j}=0\)), \(m'_j=1\), resulting in a padding mask of ones (ignore tokens).
However, a special condition applies when the entire reference input is `None': in this case, the portion of the ``key\_padding\_mask'' corresponding to the reference tokens is \emph{explicitly set to all zeros}. Consequently, the Transformer effectively \emph{only processes} the environmental observations \(o_t\), and its operational logic becomes equivalent to that of a traditional Actor-Critic network.

Once the Encoder processes the input sequence, the final representation of the \texttt{[cls]} token is fed into a Projection Head (illustrated on the right side of Fig.~\ref{fig:mmtransformer}). This head consists of a simple linear layer that maps the \texttt{[cls]} token's representation to the task-specific output. For the Actor network, this projection yields the robot's actions, a tensor of shape \((\text{num\_envs}, \text{num\_dof})\), where \(\text{num\_dof}\) is the number of degrees of freedom. For the Critic network, the \texttt{[cls]} token representation is projected to a scalar value estimate per environment, resulting in a tensor of shape \((\text{num\_envs}, 1)\). The Projection Heads for the Actor and Critic components are independent.

In our network implementation, we balance training complexity, computational cost, and memory footprint while aiming to enhance feature learning capabilities. Typically, the network employs 2-4 Transformer encoder layers. The model dimension, \(d_{\text{model}}\), is set within the range of 64-128, and 4-8 attention heads are utilized. The dimensionality of the feed-forward network (FFN) within each Transformer layer is \(4 \times d_{\text{model}}\). Notably, a configuration with \(d_{\text{model}}=64\) and 4 layers has a parameter count comparable to a standard Multi-Layer Perceptron (MLP) with layer sizes [512, 256, 128], yet it offers significantly improved representational power. The network employs GELU as the activation function. Furthermore, to mitigate potential training instability arising from large numerical disparities among different observation features, Root Mean Square Normalization (RMSNorm) \cite{zhang2019root} is applied to the sequence of input token embeddings before they are processed by the stack of Transformer encoder layers. In both dedicated reinforcement learning and subsequent imitation learning tasks, this Transformer configuration has consistently demonstrated superior optimization and generalization capabilities compared to traditional MLP architectures under equivalent conditions. We will present a detailed validation of these findings in the experimental section.

\subsection{DAgger-MMPPO Architecture and IL Integration}
\begin{figure}
    \centering
    \includegraphics[width=0.99\linewidth]{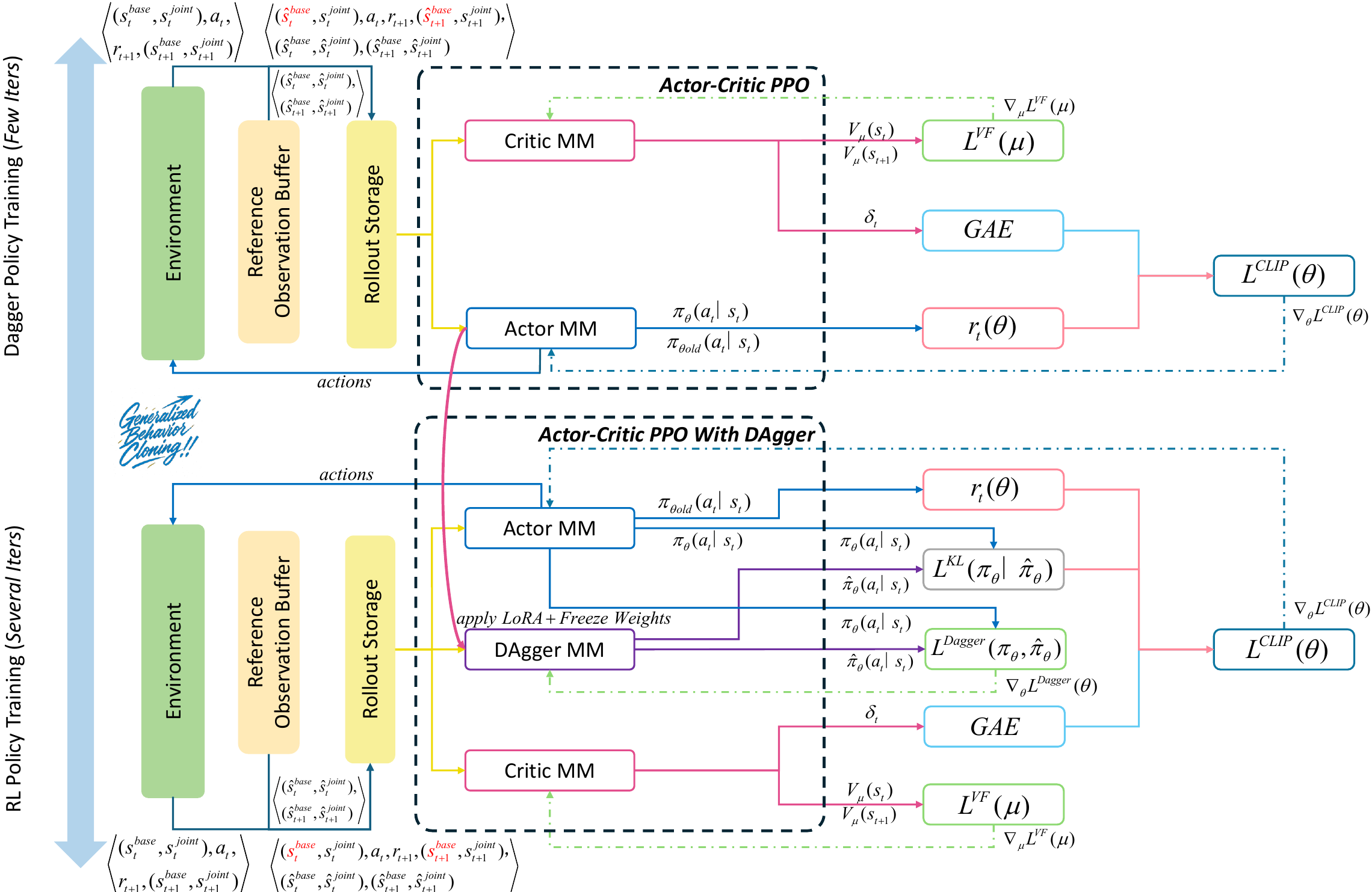}
    \caption{DAgger-MMPPO algorithm architecture. DAgger-MMPPO adopts a two stage training procedure. We first train a DAgger policy in simplified physics environment to accurately imitate human motions, then we distill the learned pattern under complex physics scene interactions.}
    \label{fig:mmppo}
\end{figure}

Although enhancements to the network backbone can improve the capacity for learning complex actions, mapping reference motions to the robot's executable actions still presents significant challenges. The agent must discern underlying patterns from a confluence of data—including joint angles, the gravity vector, reference velocities for actuators, and the positions and velocities of various links—correlate these with its own observations (\(s_t\)), and generate appropriate responses within the physical environment. Unlike direct teleoperation, this approach more closely resembles the replication of human-like motion patterns onto the robot, a process we term \emph{behavior cloning} for whole-body motion. Directly applying actions from an expert dataset is often infeasible; for instance, the balance characteristics of a referenced human pose may not be directly applicable to the robot's morphology and dynamics. To address this, on-policy reinforcement learning algorithms can be employed, enabling the agent to refine its behavior through direct interaction with the environment. The PPO algorithm is well-suited for this purpose.

We formulate the robot's behavior cloning task as a Markov Decision Process (MDP), denoted by the tuple \((\mathcal{S}, \hat{\mathcal{S}}, \mathcal{A}, R, \hat{R}, \mathcal{T}, \gamma)\). Here, \(\mathcal{S}\) represents the agent's state space, and \(\hat{\mathcal{S}}\) is the state space derived from the expert data. For an agent in state \(s_t \in \mathcal{S}\), the corresponding expert reference state \(\hat{s}_t \in \hat{\mathcal{S}}\) is retrieved from the expert dataset (e.g., based on a temporal correspondence or a mapping from \(s_t\)). \(\mathcal{A}\) is the agent's action space. \(R(s_t, a_t, s_{t+1})\) denotes the intrinsic task reward function, while \(\hat{R}((s_t, \hat{s}_t), a_t, (s_{t+1}, \hat{s}_{t+1}))\) represents the imitation reward for behavior cloning. \(\mathcal{T}(s_{t+1} | s_t, a_t)\) is the state transition probability distribution, and \(\gamma \in [0,1)\) is the discount factor for future rewards.  The term \(\lambda\) mentioned in proximity in some formulations would refer to the discount factor used in Generalized Advantage Estimation (GAE), which is relevant for algorithms like PPO. Our objective is to train a policy using the agent's observation \(o_t = \mathcal{O}(s_t)\) (where \(\mathcal{O}\) is an observation function mapping state to observation) and the corresponding reference observation \(\hat{o}_t = \mathcal{O}(\hat{s}_t)\). The policy \(\pi\) is defined as:
$$
a_t = \begin{cases}
\pi(o_t, \hat{o}_t) & \text{if } \hat{o}_t \neq \text{None}, \\
\pi(o_t) & \text{if } \hat{o}_t = \text{None}
\end{cases}
$$
This policy is trained to maximize an objective function that combines the intrinsic task reward \(R\) and the imitation reward \(\hat{R}\). However, while PPO allows the agent to continuously refine its policy through physical interaction, a significant disparity often exists between the agent's initial policy distribution \(\pi_0\), the optimal policy for the intrinsic task reward \(\bar{\pi}_R\), and the optimal policy for imitation reward \(\bar{\pi}_{\hat{R}}\). For PPO, which optimize primarily via trust region methods, finding a global optimum that satisfies these potentially conflicting objectives is challenging. This difficulty arises because the sub-optimal solutions for each objective may reside in disparate regions of the policy space, and the trust region, guiding optimization along the current steepest descent, might not encompass the path to the true optimum. Therefore, we propose \textbf{DAgger-MMPPO}, a two-stage PPO training framework that integrates concepts from the Dataset Aggregation (DAgger) algorithm with the PPO training logic to address these challenges.

\paragraph*{Stage 1: DAgger Policy Training}

\begin{algorithm}[htbp] 
\caption{Stage 1: DAgger Policy Training} 
\label{alg:mmppo_stage_1}
\begin{algorithmic}[1]
\Require 
  Expert data $\mathcal{D}_{\text{exp}}$ (provides $\hat{\mathcal{S}}$); Init. params $\theta_0, \mu_0$; Obs. func. $\mathcal{O}(\cdot)$;
  Weights $w_R, w_{\hat{R}}$ ($w_R \ll w_{\hat{R}}$); Hyperparams: $N_{\text{iter}}, K, M, B$, etc.

\State Initialize $\pi_\theta \leftarrow \theta_0, V_\mu \leftarrow \mu_0$; \\ Configure simplified env. (fixed base, prioritize $\hat{R}$).
\For{iteration $n \leftarrow 0$ \textbf{to} $N_{\text{iter}}-1$}
    \State Collect batch $\mathcal{B}=\{(s_t, a_t, R_t^{\text{task}}, \hat{s}_t, s_{t+1}, \hat{s}_{t+1}, d_t)\}$ using $\pi_\theta$.
    \State Create learning batch $\mathcal{B}_{\text{learn}}$ from $\mathcal{B}$:
    \Statex \hspace{\algorithmicindent} For each transition, replace:
    \Statex \hspace{\algorithmicindent} $s'_t \leftarrow (\hat{s}_t^{\text{base}}, s_t^{\text{joint}})$, $s'_{t+1} \leftarrow (\hat{s}_{t+1}^{\text{base}}, s_{t+1}^{\text{joint}})$.
    \Statex \hspace{\algorithmicindent} Compute $\hat{R}_t \leftarrow \hat{R}(s'_t, \hat{s}_t, a_t, s'_{t+1}, \hat{s}_{t+1})$, $R^{task}_t$.
    \Statex \hspace{\algorithmicindent} Effective reward $r_t \leftarrow w_R R_t^{\text{task}} + w_{\hat{R}} \hat{R}_t$. 
    \State Compute advantages $A_t$ and update $\mathcal{B}_{\text{learn}}$.
    \For{epoch $k \leftarrow 0$ \textbf{to} $K-1$} \Comment{PPO Update}
        \State Update $\theta, \mu$ using PPO objective on $M$ mini-batches from $\mathcal{B}_{\text{learn}}$.
    \EndFor
\EndFor
\State \Return DAgger-trained policy $\pi_D \leftarrow \pi_\theta$.
\end{algorithmic}
\end{algorithm}

The primary goal of this initial stage is to learn a \textbf{behavioral prior} that excels at pure motion imitation, temporarily decoupled from the complexities of dynamic balance. This is achieved by training an initial policy that approximates the expert policy \(\bar{\pi}_{\hat{R}}\) in a simplified physical environment (detailed in Algorithm~\ref{alg:mmppo_stage_1}). This learned policy serves two synergistic purposes: first, it provides a high-quality anchor for the full PPO exploration in Stage 2, guiding optimization from the vicinity of \(\bar{\pi}_{\hat{R}}\) towards an optimal balance with the task policy \(\bar{\pi}_R\), thus mitigating policy degradation. Second, by embedding the state-to-action mapping within the actor network, it functions as an efficient \textbf{automatic labeler}, obviating the need for computationally expensive dataset lookups.

To achieve this focused learning, we partition the agent's state \(s_t\) into \(s_t = (s_t^{\text{base}}, s_t^{\text{joint}})\). The core strategy is to \textbf{decouple the learning of joint coordination from base control}. Consequently, we impose specific environmental simplifications: the robot's base link is fixed, all physical ground interactions are disabled, and the intrinsic task reward \(R\) is minimized to bias optimization entirely towards the imitation reward \(\hat{R}\).

During training, we leverage a key insight to further isolate the joint learning problem. After collecting rollouts, we substitute the agent's observed base state \(s_t^{\text{base}}\) with the expert's reference base state \(\hat{s}_t^{\text{base}}\) as input to the policy. This substitution effectively \emph{simulates the physical characteristics of an ideally controlled base}, compelling the agent to focus solely on adjusting its joint states \(s_t^{\text{joint}}\) to maximize \(\hat{R}\). Crucially, this process forces the Stage 1 policy to learn a direct regression from the (potentially modified) state \(s_t\) to an expert-like action \(\hat{a}_t\), creating the foundational motion prior for the subsequent training phase. The update pattern is illustrated in the upper part of Figure~\ref{fig:mmppo}.

\begin{algorithm}[htbp]
\caption{Stage 2: Policy Training with DAgger}
\label{alg:mmppo_stage_2}
\begin{algorithmic}[1]
\Require DAgger-policy $\pi_D$; LoRA params $\psi_0$; Optimizers; Expert state $\hat{\mathcal{S}}$, Obs. func. $\mathcal{O}(\cdot)$;
 $w_{\text{im}}$ (imit. weight); $\rho$ (DAgger factor, anneals); PPO Hyperparameters ($\gamma, \lambda, N_{\text{iter}}, K, M, B, \epsilon_{\text{clip}}$).

\State Init actor $\pi_\theta \leftarrow \pi_D$, critic $V_\mu\leftarrow \mu_0$. Init DAgger net $\pi_{\text{Dagg}}$ (frozen $\pi_D$, LoRA $\psi \leftarrow \psi_0$).
\State Configure full phys. env.; Define task reward $\bar{R}_t=(R_t, \hat{R}_t)$.

\For{iteration $n \leftarrow 0$ \textbf{to} $N_{\text{iter}}-1$}
    \State Collect batch $\mathcal{B}$ via $\pi_\theta(o_t, \hat{o}_t)$ using $\bar{R}_t$.
    \State Compute advantages $A_t$ for $\mathcal{B}$ using $\bar{R}_t, V_\mu(o_t, \hat{o}_t)$, GAE($\gamma, \lambda$).
    \For{epoch $k \leftarrow 0$ \textbf{to} $K-1$} \Comment{DAgger-MMPPO Update}
        \For{mini-batch from $\mathcal{B}$ (shuffled)}
            \State Compute $L^{\text{CLIP}}(\theta)$ for actor $\pi_\theta$.
            \State Generate $\hat{a}_m^{\text{Dagg}} \leftarrow \pi_{\text{Dagg}}(o,\hat{o})$.
            \State Compute $L^{\text{KL}}(\theta)$ using $\hat{a}_m^{\text{Dagg}}$.\Comment{(Eq.~\ref{eq:im_loss})}
            \State Update $L^{\text{CLIP}}(\theta)\leftarrow L^{\text{CLIP}}(\theta) + w_{\text{im}} L^{\text{KL}}(\theta)$.
            \State Compute $L^{\text{VF}}(\mu)$ for critic $V_\mu$.
            \State Update actor $\theta$ using $\nabla_L^{\text{CLIP}}(\theta)$; Update critic $\mu$ using $\nabla_\mu L^{\text{VF}}(\mu)$.
            
            \State Compute $L^{\text{DAgger}}_{\psi}(a_m^{\text{Actor}}, \hat{a}_m^{\text{Dagg}}, \rho)$. \Comment{(Eq.~\ref{eq:dagger_loss})}
            \State Update LoRA params $\psi$ using $\nabla_\psi L_{\text{DAgger}}(\psi)$.
        \EndFor
    \EndFor
    \State Anneal $\rho$.
\EndFor
\State \Return Trained policy $\pi_\theta$.
\end{algorithmic}
\end{algorithm}

\paragraph*{Stage 2: Policy Training with DAgger}
Although the initial policy \(\pi_D\) from Stage 1 is proficient, its learned \textbf{dynamics prior} is conditioned on a simplified physical environment, rendering it non-deployable in a full-physics simulation. Stage 2 is therefore designed to \textbf{adapt this expert prior} to the task-optimal policy \(\bar{\pi}_R\) that is robust to complex physical interactions. To achieve this, we augment PPO with a DAgger-like supervisory component. This process can be viewed as a sophisticated variant of \textbf{teacher-student distillation} \cite{hinton2015distilling,rusu2015policy,tang2019distilling}, where the teacher provides a powerful but imperfect control prior, and the student learns to make it viable in a more complex reality. This leads to our DAgger-MMPPO core training framework (Fig.~\ref{fig:mmppo}, lower part; Algorithm~\ref{alg:mmppo_stage_2}).

In this training phase, we instantiate two networks from the prior \(\pi_D\), reminiscent of an \textbf{elite simulator pilot (the "teacher")} and a \textbf{real-world pilot (the "student")}.
\begin{itemize}
    \item The \textbf{DAgger network serves as the "teacher"}. It embodies the dynamics prior learned in the simplified Stage 1 environment. It provides expert control actions that are highly effective in its idealized world, but do not account for real-world complexities like balance or ground reaction forces.
    \item The main \textbf{actor network acts as the "student"}. Its objective is to learn a physically robust policy via PPO. It takes the teacher's suggested actions as a strong reference but must learn to adapt them to succeed in the full-dynamics environment.
\end{itemize}
To ensure the teacher's guidance remains relevant as the student explores, we employ \textbf{Low-Rank Adaptation (LoRA)} \cite{hu2022lora} to subtly fine-tune the teacher network based on the student's interaction data. This creates a symbiotic learning process: the student (actor actions \(a_t\)) learns to ground the teacher's idealized strategy in reality, while the teacher (DAgger predictions \(\hat{a}_t\)) is concurrently refined, providing increasingly pertinent guidance.

For actor optimization, we consider the discrepancy between its policy \(\pi_{\theta}\) and the DAgger network's output \(\pi_D\). Direct KL divergence calculation is time-consuming as \(\pi_D\) is also needs to be sampled from its distribution. Inspired by imitation loss design in \cite{yao2024anybipeendtoendframeworktraining}, we treat the DAgger network's output action \(\hat{a}_t\) as the mean of a Dirac delta distribution. This \(\hat{a}_t\) serves as a constant target (no gradients to DAgger network during actor optimization). The actor's imitation loss is the negative log-likelihood of these target actions under its policy \(\pi_\theta(\cdot|s_i)\):
\begin{equation}
\begin{aligned}
    L_{\text{KL}}(\pi_\theta\mid \hat{\pi}_\theta) &= \mathbb{E}_{s_t \sim \mathcal{D}} \left[ -\log \pi_\theta(\hat{a}_t | s_t) \right] \\
    &\approx \frac{1}{N} \sum_{i=1}^{N} \Biggl[ \log\left(\sqrt{2\pi} \sigma_{\theta}(s_i)\right) \\
    &\qquad\qquad + \frac{(\hat{a}_i - \mu_{\theta}(s_i))^2}{2\sigma_{\theta}(s_i)^2} \Biggr]
\end{aligned}
\label{eq:im_loss}
\end{equation}

where \(\hat{a}_i = \text{DAgger}(s_i)\), and \(\mu_{\theta}(s_i), \sigma_{\theta}(s_i)\) are the actor's Gaussian policy outputs. The expectation is over a batch of \(N\) states.

For DAgger network LoRA fine-tuning, a hyperparameter \(\rho \in [\rho_{\text{min}}, \rho_{\text{max}}] \subset [0,1]\) anneals over iterations. This \(\rho\) balances DAgger network's self-consistency against adaptation to the actor. The DAgger network (outputting \(\hat{a}_t^{\text{Dagg}}\)) is guided by actor's actions \(a_t^{\text{Actor}}\). The DAgger loss is mainly:
\begin{equation}
L_{\text{DAgger}}(\hat{a}_t^{\text{Dagg}}, a_t^{\text{Actor}}, \rho) = (1-\rho) \|\hat{a}_t^{\text{Dagg}} - a_t^{\text{Actor}}\|_2^2
\label{eq:dagger_loss}
\end{equation}
This loss aligns the DAgger network's output \(\hat{a}_t^{\text{Dagg}}\) with the actor's \(a_t^{\text{Actor}}\). The alignment strength is scaled by \((1-\rho)\); as \(\rho\) decreases, the pull towards the actor's policy increases. To ensure stability and prevent the DAgger network from being excessively influenced by the actor, its LoRA updates are implicitly regularized by considerations similar to a GAN-like gradient penalty, supplementing explicit measures like gradient clipping. Thus, for low \(\rho\), the overall DAgger network objective is considered minimized when its output \(\hat{a}_t^{\text{Dagg}}\) closely matches \(a_t^{\text{Actor}}\) (minimizing the \(L_{\text{DAgger}}\) term above) and the network satisfies these stability criteria (e.g., associated penalized gradients are small).

Our framework's modular design allows for seamless integration with established imitation learning algorithms to enhance performance. For instance, incorporating \textbf{Adversarial Motion Priors (AMP)} can reduce reliance on meticulously engineered reward functions like \(\hat{R}\). Our data formulation, which includes the reference states \(\hat{S}\), naturally provides the consecutive state-transition pairs (\((o_t, o_{t+1})\) and \((\hat{o}_t, \hat{o}_{t+1})\)) required to train an AMP discriminator on motion style, with a masking mechanism to handle environments lacking reference data. Additionally, we introduce an auxiliary \textbf{symmetry loss}. Leveraging the symmetric data augmentation from our pipeline (Section~\ref{subsec:pipeline}), this loss encourages the PPO policy to produce appropriately mirrored actions for mirrored inputs, promoting behavioral symmetry. Crucially, these optimizations are integrated directly into the main RL training loop, obviating the need for complex reward engineering or robot-specific adaptations. This demonstrates the flexibility of our framework\footnote{Our DAgger-MMPPO algorithm library is adapted from the open-source rsl\_rl project and is available at \url{https://github.com/sjtu-mvasl-robotics/rsl_rl}}, which is compatible with recent versions of Isaac Lab (tested as of June 2025).

\section{Training Platform Design}
\label{sec:system}
To facilitate seamless integration with contemporary robotic reinforcement learning ecosystems, we have adopted Isaac Lab as our foundational platform. Compared to its predecessor, Isaac Gym (which is currently deprecated by NVIDIA), Isaac Lab employs a Manager-Based software architecture. This architecture encapsulates core functionalities—such as the foundational training environment, physics simulation, and data management utilities—into readily accessible interfaces. Consequently, unlike its predecessor, it obviates the need to design bespoke Gym environments and custom environment-specific computational logic for distinct robot configurations. Instead, users primarily need to configure robot-specific model parameters, observation spaces, and reward functions. The adoption of this framework significantly streamlines the process of ensuring compatibility across diverse robot configurations within our design. This section will detail the foundational system design of our GBC framework, our approach to curriculum learning\cite{bengio2009curriculum} and the design of imitation reward functions, and specific optimizations implemented for imitation training.

\subsection{RL Platform Design}
\label{subsec:rl_platform_design}
\begin{figure}[htbp]
    \centering
    \includegraphics[width=0.98\linewidth]{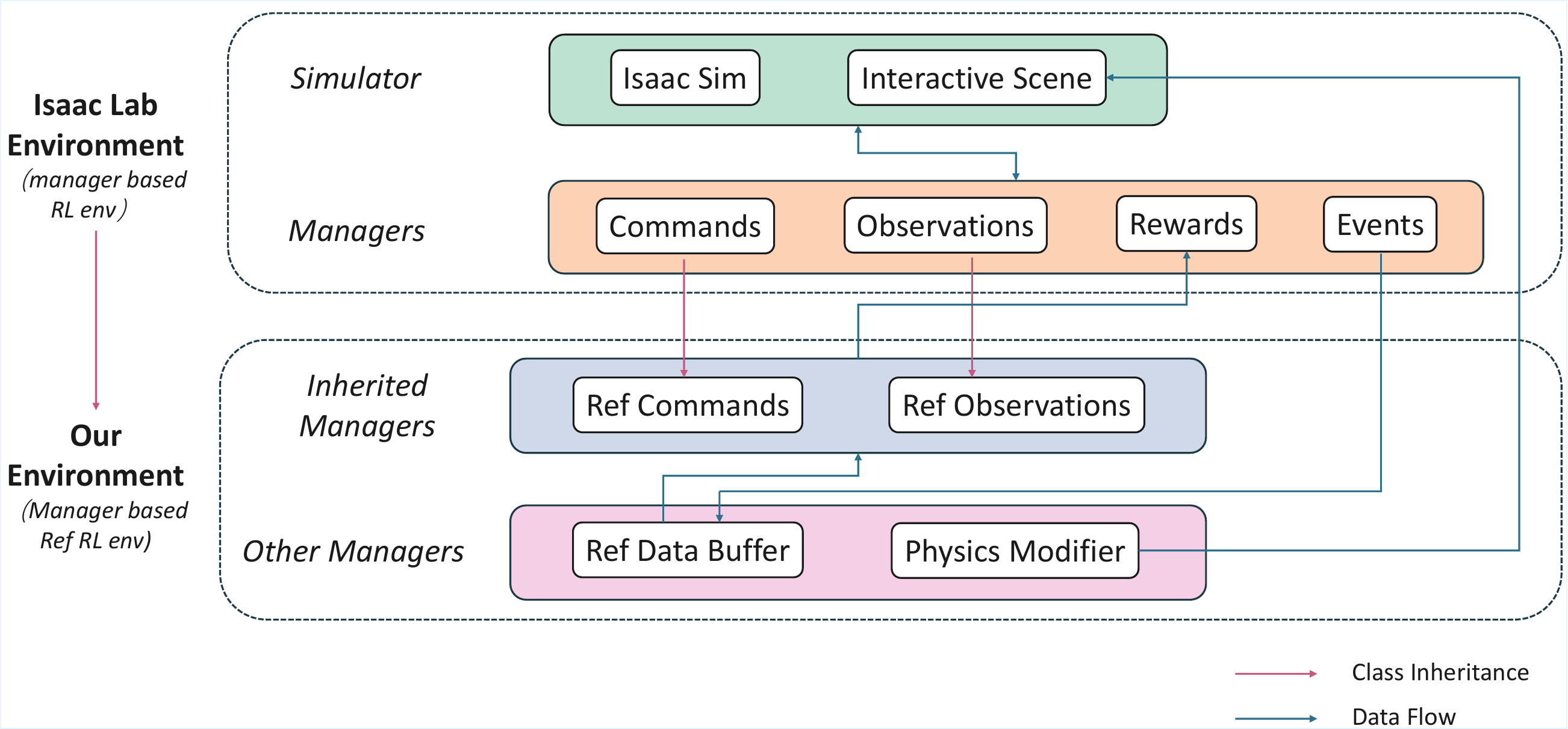}
    \caption{GBC platform relationship with Isaac Lab Environments. We inherited the manager design from official base and specialized to be compatible with reference data. We also created some managers of our own to support curriculum learning and imitation training.}
    \label{fig:system-design}
\end{figure}

Isaac Lab features a decoupled and encapsulated architecture for reinforcement learning environments, which allows us to integrate most functionalities by inheriting from its base environment classes. Figure~\ref{fig:system-design} illustrates the additional managers we introduced and their inheritance relationships within the existing framework. To achieve efficient loading and accessing of reference datasets, we implemented a \emph{RefDataBuffer} responsible for caching transformed robot reference motion datasets onto the GPU. This class maintains a GPU-resident buffer, storing individual motion sequences and managing their corresponding indices for each reinforcement learning environment. It computes the appropriate reference data index based on the current RL environment's timestep and incorporates support for cyclic data indexing, thereby enabling low memory footprint yet efficient retrieval. As all cached data primarily consists of differential quantities (e.g., velocities, accelerations), the class also implements efficient on-the-fly integration to derive absolute states, accelerated using CUDA \emph{warp-level primitives} where applicable. This overall design ensures \(O(1)\) query complexity for direct data retrieval. Furthermore, the integration process to obtain absolute states typically maintains \(O(1)\) complexity by leveraging caching mechanisms and localized (e.g., neighboring point) lookups. Only in infrequent worst-case scenarios, such as an environment reset requiring a full re-integration from the beginning of a sequence, does this operation degrade to \(O(N)\), which is itself optimized for parallel execution.

Analogous to Isaac Lab's standard \emph{ObservationManager}, we implemented a \emph{ReferenceObservationManager} dedicated to managing the querying and updating of reference states \(\hat{s}_t\). This manager is invoked at the beginning of each simulation step to refresh the reference state buffer for the current timestep, ensuring \(O(1)\) access latency for components such as the \emph{CommandGenerator} and reward functions. Furthermore, this manager curates the reference observations \(\hat{o}_t\) and their corresponding availability masks based on the presence or accessibility of reference data in the current context. Users can load required reference data by specifying \emph{term\_name} as saved in the dataset. The manager automatically handles symmetry transformations for data augmentation, based on the properties of these terms, such as their relation to spatial angles (e.g., angular velocities) or positions (e.g., coordinates, linear velocities). We also allow users to selectively configure whether each term contributes to the reference observation \(\hat{o}_t\). Terms not designated as observations remain invisible to the PPO network but can still be utilized in reward computations.

To ensure consistency between available reference information and the actions commanded during training, we have overridden the default \emph{CommandManager}. When reference state data is available for the current environment, our custom \emph{CommandManager} replaces the default stochastic command generation (e.g., for target linear velocities \(v_x, v_y\) and angular velocity \(\omega_z\)) with target velocities derived from the reference motion \(\hat{s}_t\). This modification is intended to assist the agent in learning the mapping from reference velocities to desired actions during training.

Our framework facilitates the rapid extension of existing Isaac Lab training tasks into imitation learning scenarios. Users can inherit an environment's baseline observation definitions and physical descriptions. Subsequently, they can integrate imitation-specific configurations—such as reference state specifications, imitation reward functions, AMP networks, and symmetry settings—through our provided imitation learning library, thereby streamlining the setup of new imitation learning experiments.

\subsection{Curriculum Learning and Imitation Reward Design}
Directly imitating precise joint configurations and end-effector positions can be challenging for an agent, potentially leading to sparse rewards during initial exploration. This sparsity can impede the convergence speed and motivation towards accurate imitation. We therefore employ curriculum learning, enabling the agent to commence exploration in simpler scenarios and progressively increasing the training difficulty as its imitation proficiency improves. To this end, we have designed two distinct curricula based on reward function modulation and environmental modifications: a \emph{standard-variation curriculum} and a \emph{toddler curriculum}, which implement staged learning through reward shaping and interactive assistance, respectively.

\paragraph*{Standard-variation curriculum}
For this curriculum, we define the imitation reward for each relevant term using an exponential function. The reward for a specific term \(s^{\text{term}}_t\) in relation to its target \(\hat{s}^{\text{term}}_t\) is given by:
\begin{equation}
R^{\text{term}}_t = \omega^{\text{term}} \exp\left(-\frac{\|s^{\text{term}}_t - \hat{s}^{\text{term}}_t\|^2}{2(\sigma^{\text{term}}_l)^2}\right)
\label{eq:exp_reward_corrected}
\end{equation}
where \(\omega^{\text{term}}\) is a weighting factor, \(s^{\text{term}}_t\) is the agent's current state component, \(\hat{s}^{\text{term}}_t\) is the corresponding expert target, and \(\sigma^{\text{term}}_l\) is the standard deviation for that term at curriculum level \(l\). (Note: The original formula provided in a previous prompt, \( (s^{\text{term}}_t-\hat{s}^{\text{term}}_t) \), has been interpreted as the squared L2-norm \( \|s^{\text{term}}_t - \hat{s}^{\text{term}}_t\|^2 \) in the numerator, which is conventional for such exponential reward structures. If a scalar, non-squared difference is strictly intended, the formula should be revised.) This reward formulation possesses favorable properties: ignoring the multiplier \(\omega^{\text{term}}\), the function's range is \([0,1]\), and a larger \(\sigma^{\text{term}}_l\) facilitates easier attainment of higher rewards. Users can specify an initial \(\sigma^{\text{term}}_0\) and a schedule for its gradual reduction. A specific update threshold, \(\text{ratio} \in (0,1)\), is also defined. When the normalized reward \(R^{\text{term}}_t / \omega^{\text{term}}\) consistently meets or exceeds this ratio for a defined period, the standard deviation \(\sigma^{\text{term}}_l\) is reduced, making the training conditions more stringent. Initially, these terms typically have a large \(\sigma\), aiding rapid exploration towards a convergent direction. To sustain motivation for further refinement and prevent premature convergence to local optima, the difficulty of achieving rewards is progressively increased as the agent's state approaches the reference state.

\paragraph*{Toddler curriculum}
To facilitate smoother convergence of the policy from the DAgger-learned distribution \(\pi_D\) to behavior suitable for the complexities of the physical environment, particularly during initial interactions involving contact, we introduce a "toddler curriculum." Inspired by a toddler's walker, this curriculum provides assistive forces. Specifically, when the robot's base height \(h^{\text{base}}\) falls below a reference height threshold (\(\hat{h}^{\text{base}} + \text{offset}\)), a spring-like force is applied solely in the vertical (z) direction (this force is active only when \(h^{\text{base}} < \hat{h}^{\text{base}} + \text{offset}\)):
\begin{equation}
\begin{aligned}
    F_z^{\text{base}} =& -k \cdot \text{clamp}\left((h^{\text{base}}-(\hat{h}^{\text{base}}+\text{offset})), -h_{\text{max}}, 0\right)\\& - c \cdot \dot{h}^{\text{base}}
\end{aligned}
\label{eq:toddler_force_corrected}
\end{equation}

Alternatively, a strong constraining force could be applied if the height deviates significantly. To prevent unnatural accelerations, the maximum assistive spring force (derived from \(k \cdot h_{\text{max}}\)) is bounded. Throughout learning, as imitation reward improves and falls decrease, the parameters \(k\) and the effective assistance range (e.g., \(h_{\text{max}}\)) are gradually reduced until the assistive force is nullified.

The advantage of this toddler curriculum is the circumvention of complex physical models like suspension rigs. The assistive force acts only along the global z-axis, minimizing interference with other imitated behaviors. This strategy primarily aims to prevent early-stage falls due to suboptimal dynamic interactions, effectively allowing the agent to "first learn a stable walking posture, then refine the gait." The robot will still fall if its actions deviate excessively, ensuring termination conditions remain effective even with initially imperfect actions. This functionality is implemented by an additional \emph{PhysicsModifierManager} within our Isaac Lab framework, defining a scheme for dynamically adjusting external forces. This manager's extensible design also supports other dynamic physics modifications, such as applying varying forces to a robot's arm during manipulation tasks.

The imitation reward function, \(\hat{R}\), is composed of several terms summarized in Table~\ref{tab:imitation_rewards}. These terms are categorized by their mathematical formulation and the specific state components they target. For conciseness, the primary task reward components, \(R\), used for general robot control are omitted from this summary.

\begin{table}[htbp]
\centering
\begin{threeparttable}
\caption{Summary of imitation reward (\(\hat{R}\)) components.}
\label{tab:imitation_rewards}
\begin{tabular}{@{} >{\raggedright\arraybackslash}p{2.2cm} 
                     >{\raggedright\arraybackslash}p{2.8cm} 
                     >{\raggedright\arraybackslash}p{2.6cm} @{}} 
\toprule
Reward Type & Reward Terms (Examples) & Expression \\
\midrule
Exponential Penalty (with curriculum \(\sigma_l\)) & Joint positions (raw or normalized), Joint velocities, Root linear/angular velocities & \( \omega \exp(-\tfrac{\|s_t - \hat{s}_t\|^2}{2\sigma_l^2}) \) \\
\addlinespace
Exponential Penalty (without curriculum) & Root global position/velocity, Link global positions/velocities, Target height & \( \omega \exp(-\tfrac{\|s_t - \hat{s}_t\|^2}{2\sigma^2}) \) \\
\addlinespace
L2 Norm Deviation & Gravity orientation, Target end-effector orientation & \( -\omega \|s_t - \hat{s}_t\|^2 \) \\
\addlinespace
L2 Norm of Rate of Deviation & Joint accelerations, Root tracking deviation rate & \( -\omega \| \dot{s}_t - \dot{\hat{s}}_t \|^2 \)~\tnote{a} \\
\addlinespace
Binary Match Sum (per frame) & Feet in contact (per foot, binary state) & \( \omega \neg (s_t\oplus \hat{s}_t) \)~\tnote{b} \\
\bottomrule
\end{tabular}
\begin{tablenotes}
    \item[a] \(\dot{s}_t\) denotes the time derivative of \(s_t\).
    \item[b] \(\oplus\) is the XOR operator, since contact indices are binary.
\end{tablenotes}
\end{threeparttable}
\end{table}

Given the incorporation of a symmetry loss term directly into the PPO objective (as discussed previously), the design of these imitation reward components does not require explicit balancing for the tracking of symmetric body parts; the policy network is expected to learn appropriate mirrored responses inherently.

\subsection{Training Optimizations}

To enhance learning efficiency and policy generalization, we incorporate several critical optimization strategies during training. These strategies are designed to create a robust and varied training distribution.

\textbf{Domain and Reference Randomization.}
To bridge the sim-to-real gap and improve robustness, we apply extensive domain randomization. This includes randomizing the robot's physical properties (e.g., payload mass, friction), actuator parameters, and applying external pushes. We also inject noise into both agent and reference observations to encourage the policy to master motions even from imperfect demonstrations \cite{brown2020better}.

\textbf{Randomized Start-State Initialization.}
Following the principle from DeepMimic \cite{2018-TOG-deepMimic}, we randomize the starting point within each motion sequence at the beginning of every episode. Consistently starting from a fixed point can bias the policy, whereas randomizing the initial state encourages the agent to learn robust recovery strategies from any point in a trajectory. Our \emph{RefDataBuffer} class (detailed in Subsection~\ref{subsec:rl_platform_design}) facilitates this by efficiently computing the robot's initial global pose, joint states, and velocities from stored differential motion data, ensuring correct physical initialization even for dynamic states like jumping or rolling.

\textbf{Randomized Motion Selection.}
To prevent overfitting to a small set of motions, we also randomize the reference motion sequence assigned to each environment upon reset. Unlike some frameworks that improve stability by tracking a single sequence per training round \cite{fu2024humanplus, he2024learning}, our approach exposes the policy to a diverse set of motions from datasets like AMASS. This substantially enhances the policy's generalization capabilities, which we empirically observed in cross-dataset transfer tasks.

\textbf{Imitation-Aware Early Termination.}
Standard termination conditions, such as resetting on base contact, are insufficient for whole-body imitation, as motions like crawling require ground contact. To prevent the agent from learning non-imitative, survival-prolonging behaviors (e.g., "push-ups"), we introduce stricter, imitation-aware termination conditions. An episode terminates if the agent's state deviates significantly from the reference, such as a large discrepancy in base orientation relative to gravity or in root height. This ensures the policy remains closely aligned with the demonstration.

\textbf{Auxiliary Learning Objectives.}
Finally, we integrate supplementary reinforcement strategies. To promote symmetric behaviors, we augment observations with a predefined symmetric mapping and add an auxiliary symmetry loss, conceptually similar to Equation~\ref{eq:sym}, to the PPO objective. For methods like AMP, we also randomize the sampling of motion segments to ensure a uniform data distribution for the discriminator.

\section{Experiments}

\subsection{Overview of Experiments}
As a comprehensive framework for imitation learning, the Generalized Behavior-Cloning (GBC) framework consists of several interconnected modules. This section outlines the structure of our empirical evaluation, which is designed to rigorously validate each component and the framework as a whole. We will present a series of targeted experiments, each with its own specific setup and metrics, conducted across the \textbf{NVIDIA Isaac Sim}\cite{NVIDIA_Isaac_Sim} and \textbf{MuJoCo} \cite{todorov2012mujoco} simulation platforms. The experimental validation is organized as follows:

\begin{figure}[htbp]
    \centering
    \includegraphics[width=0.96\linewidth]{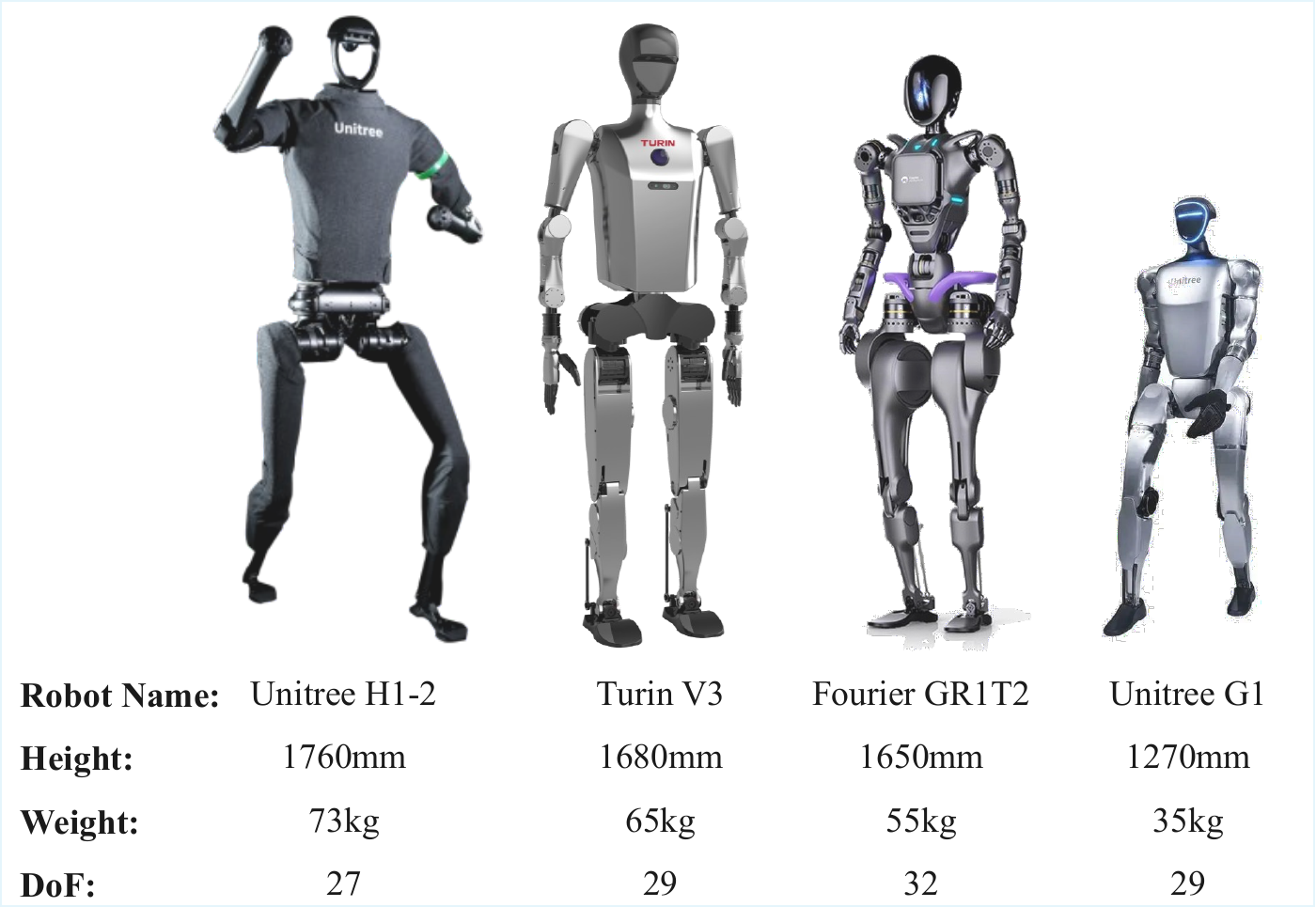}
    \caption{Robot names and parameters used in the experiment part.}
    \label{fig:robots_used}
\end{figure}

\paragraph{Data Processing Pipeline Validation}
The first part of our evaluation focuses on the data pipeline. We will demonstrate the efficacy of our method across four distinct humanoid morphologies shown in Fig.~\ref{fig:robots_used}. This includes showcasing the quality of the initial SMPL+H model fitting, the performance of the trained pose retargeting network, the smoothing effects of our post-processing routine, and the final executable quality of the generated motions in Isaac Sim. Crucially, we will also show the retargeting network's powerful generalization capability on motions not present in its training set. The resulting convergence metrics can serve as a practical benchmark for others implementing similar retargeting networks.

\paragraph{MMTransformer Backbone Analysis}
Next, we analyze the performance of our proposed MMTransformer backbone. For specific reinforcement learning (RL) and imitation learning (IL) tasks on a given humanoid, we will conduct a comparative analysis against a standard MLP backbone, focusing on parameter efficiency, training convergence profiles, and final policy performance. Furthermore, we will present an ablation study on the MMTransformer's internal design, comparing the convergence and effectiveness of different observation embedding architectures. We will also compare it with standard AMP+PPO setting to prove our architecture outperforms conventional reward-based imitation learning methods.

\paragraph{DAgger-MMPPO Algorithm Evaluation}
The third part evaluates the effectiveness of the complete DAgger-MMPPO algorithm. Using motion sub-datasets from AMASS, we will benchmark the full GBC framework with its internal ablation (our MMPPO policy trained with different algorithm configuration groups). Within this section, we will also isolate and quantify the contributions of key components---namely the DAgger part and the Toddler Curriculum---to the improvements in training speed, final performance, and overall stability on a representative task.

\paragraph{Generalization and Transferability Assessment}
Finally, we assess the generalization and transfer learning capabilities of the policies trained with GBC. We will evaluate the policy's ability to track novel motions from held-out AMASS test sets and from newly converted human motion data. Additionally, we will examine the sim-to-sim transferability of the learned policies by deploying them across different simulation environments to verify their robustness to variations in physics and dynamics.

\subsection{Data Processing Pipeline Validation}
In this subsection, we present the empirical validation of our data processing pipeline. We begin by outlining the fundamental configuration for motion retargeting, including the training setup and the metrics used for evaluation.

A critical first step is to establish a correspondence between the SMPL+H body model joints and the respective links of each target humanoid, which is essential for accurate pose matching. For each correspondence, we assign a \texttt{joint\_weight} that determines its importance during the shape-fitting and retargeting optimization processes. Assigning a higher weight to end-effectors, for instance, helps ensure that their final positions are more precise and physically plausible. The detailed mappings for the four humanoid platforms are provided in Table~\ref{tab:joint_mapping}.

\begin{table*}[t]
\centering
\caption{Correspondence of SMPL+H Joints to Robot Links and Assigned Weights}
\label{tab:joint_mapping}
\begin{tabular}{@{}l l l l l c@{}}
\toprule
\textbf{Joint Name} & \textbf{Link Name (H1-2)} & \textbf{Link Name (Turin V3)} & \textbf{Link Name (Fourier GR1)} & \textbf{Link Name (Unitree G1)} & \textbf{Joint Weight} \\
\midrule
Pelvis & pelvis & base\_link & base\_link & pelvis & 1.0 \\
.*\_Hip & .*hip\_yaw\_link & .*hip\_yaw\_link & .*thigh\_pitch\_link & .*hip\_roll\_link & 1.0 \\
.*\_Knee & .*knee\_link & .*knee\_link & .*shank\_pitch\_link & .*knee\_link & 1.5 \\
.*\_Ankle & .*ankle\_pitch\_link & .*ankle\_roll\_link & .*foot\_roll\_link & .*ankle\_roll\_link & 4.0 \\
.*\_Toe & - & .*append\_toe\_link\textsuperscript{*} & - & - & 4.0 \\
.*\_Shoulder & .*shoulder\_pitch\_link & .*arm\_roll\_link & .*upper\_arm\_roll\_link & .*shoulder\_roll\_link & 1.0 \\
.*\_Elbow & .*elbow\_link & .*elbow\_pitch\_link & .*lower\_arm\_pitch\_link & .*elbow\_link & 1.5 \\
.*\_Wrist & .*wrist\_pitch\_link & .*wrist\_pitch\_link & .*hand\_roll\_link & .*wrist\_yaw\_link & 2.0 \\
\bottomrule
\multicolumn{6}{l}{\textsuperscript{*}\footnotesize{For the Turin V3, we added an additional fixed joint and a virtual link to improve regression for foot pitch angles.}}
\end{tabular}
\end{table*}

Using these mappings, we perform the initial shape-fitting by optimizing the SMPL+H shape parameters to align with the robot's morphology in both its default joint configuration and a standard T-pose. The high-precision results of this fitting process are summarized in Table~\ref{tab:shape_fitting_results}. These results confirm that with appropriate joint correspondences, our method achieves a highly accurate mapping from the human model to the robot's kinematic structure. This creates a solid foundation for the subsequent, more detailed joint retargeting stage, enabling any motion defined under the SMPL+H standard to be effectively translated into the robot's action space.

\begin{table}[h]
\centering
\caption{SMPL+H Shape Fitting Results}
\label{tab:shape_fitting_results}
\begin{tabular}{@{}l c c c@{}}
\toprule
\textbf{Robot Name} & \textbf{Max Pos. E (cm)} & \textbf{Avg. Pos. E (cm)} & \textbf{Avg. Pose Loss} \\
\midrule
Unitree H1-2 & 1.73 & 0.86 & 0.074 \\
Turin V3 & 1.86 & 0.91 & 0.082 \\
Fourier GR1 & 1.18 & 0.47 & 0.040 \\
Unitree G1 & 1.06 & 0.44 & 0.038 \\
\bottomrule
\end{tabular}
\end{table}

For the Pose Transformer retargeting network, we sampled the entire AMASS dataset at 10-frame intervals, splitting the data into an 80:20 train/validation set. We used data regressed from the HumanAct12 dataset~\cite{guo2020action2motion} as an out-of-distribution test set. The loss weights for $\mathcal{L}_{dist}$, $\mathcal{L}_{limit}$, $\mathcal{L}_{disturb}$, and $\mathcal{L}_{sym}$ (as defined in Section III-A) were set to 5000.0, 1000.0, 100.0, and 1000.0, respectively. To prevent the regularization term from hindering initial optimization, the weight for the Action Disturbance Loss was annealed from 100.0 to 200.0 during training. All models were trained for 100 epochs with a learning rate of 1e-4. The final retargeting accuracy is detailed in Table~\ref{tab:pose_transformer_results}.

\begin{table*}[t]
\centering
\caption{Pose Transformer Retargeting Accuracy (End-Effector Position Error, in meters)}
\label{tab:pose_transformer_results}
\begin{tabular}{@{}l ccc ccc ccc@{}}
\toprule
{\textbf{Robot Name}} & \multicolumn{3}{c}{\textbf{Max Position Error}} & \multicolumn{3}{c}{\textbf{Average Position Error}} & \multicolumn{3}{c}{\textbf{Median Position Error}} \\
\cmidrule(lr){2-4} \cmidrule(lr){5-7} \cmidrule(lr){8-10}
 & Train & Val & Test & Train & Val & Test & Train & Val & Test \\
\midrule
Unitree H1-2 & 0.4367 & 0.4305 & 0.4313 & 0.0563 & 0.0555 & 0.0548 & 0.0251 & 0.0232 & 0.0237 \\
Turin V3 & 0.4486 & 0.4471 & 0.4464 & 0.0581 & 0.0565 & 0.0572 & 0.0266 & 0.0248 & 0.0254 \\
Fourier GR1 & 0.2864 & 0.2746 & 0.2562 & 0.0476 & 0.0464 & 0.0428 & 0.0164 & 0.0156 & 0.0158 \\
Unitree G1 & 0.2487 & 0.2324 & 0.2268 & 0.0432 & 0.0418 & 0.0416 & 0.0157 & 0.0144 & 0.0142 \\
\bottomrule
\end{tabular}
\end{table*}

\begin{figure*}[hbt!]
    \centering
  \subfigure[Main Loss for H1-2.]{%
    \includegraphics[width=0.4\linewidth]{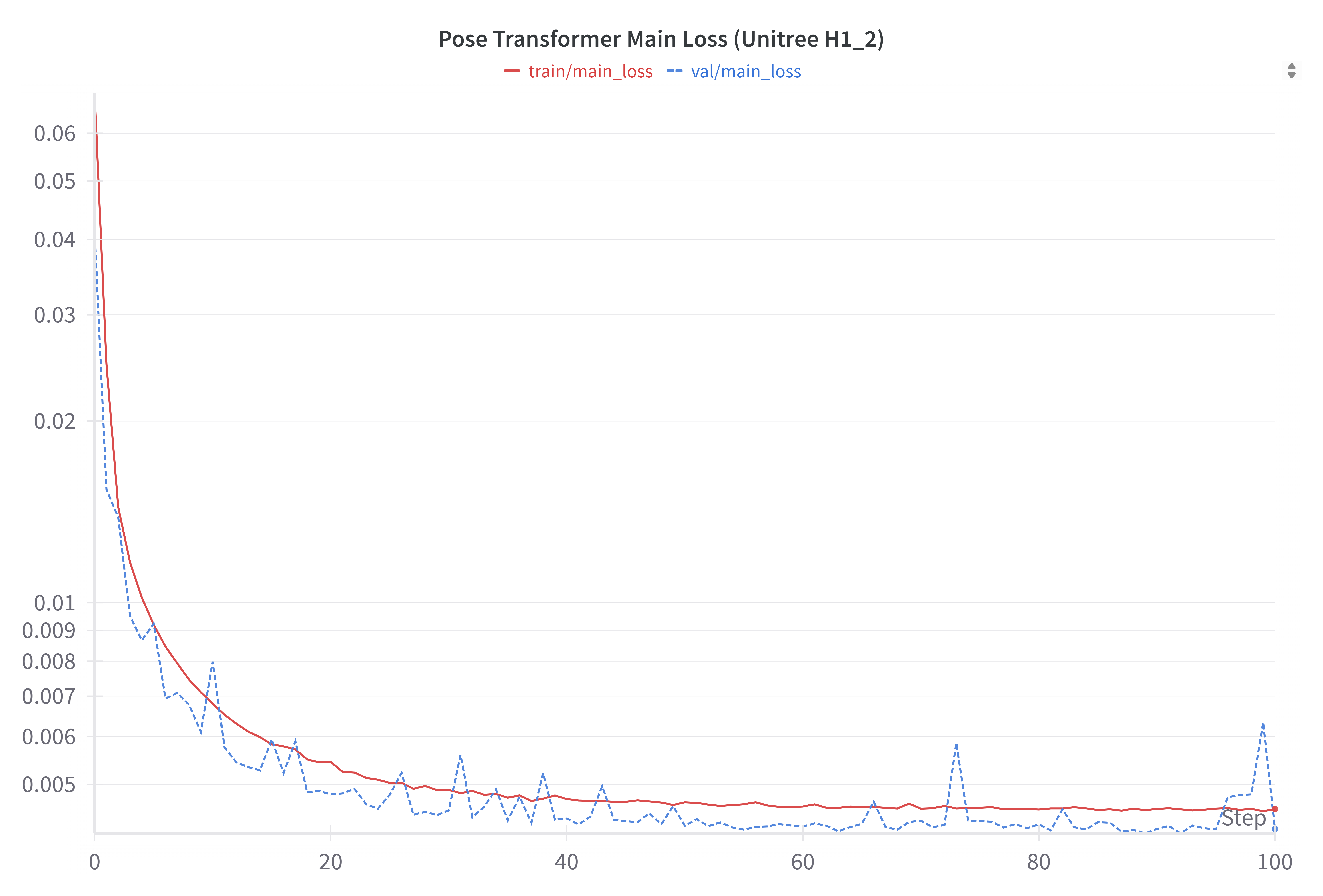}}
  \hfill
  \subfigure[Main Loss for GR1.]{%
    \includegraphics[width=0.4\linewidth]{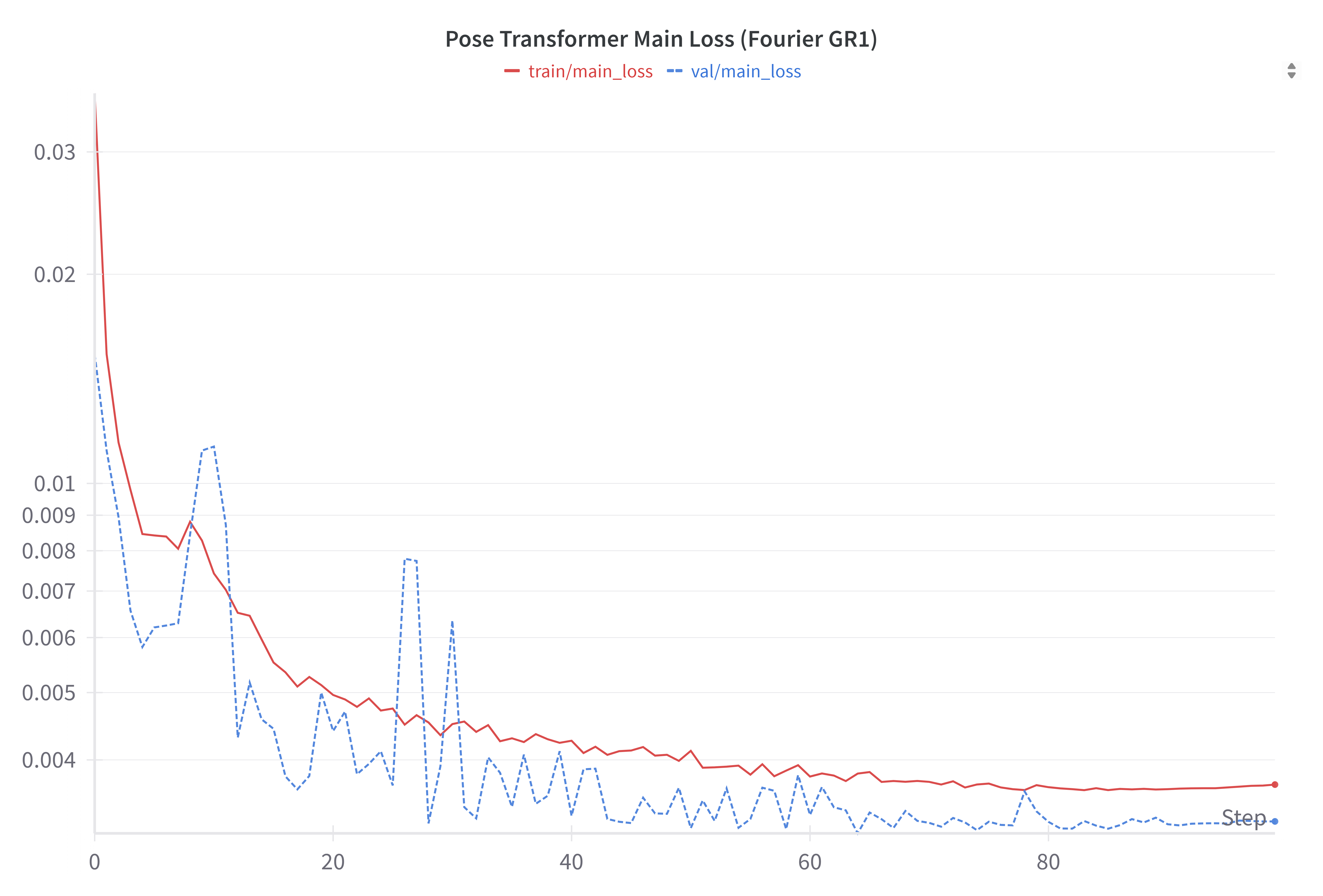}}
    
    \caption{Training and validation loss curves for the Pose Transformer network on the Unitree H1-2 and Fourier GR1 robots. The horizontal axis represents the training epochs, while the vertical axis represents the total loss value. The close proximity and stable convergence of both the training loss (blue) and validation loss (orange) indicate that the network did not overfit to the training data and exhibits strong generalization to unseen validation samples.}
    \label{fig:main_loss_gr1_h1}
\end{figure*}

The results show that the Fourier GR1 and Unitree G1 models, which feature a torso pitch joint, achieve better fitting accuracy and thus lower maximum error on motions involving torso movement. In contrast, the H1-2 and Turin V3, lacking this DoF, exhibit slightly higher maximum errors. A significant portion of the AMASS dataset contains motions that are not fully kinematically reachable by the robots. In these cases, the Pose Transformer provides a close approximation within the robot's physically feasible action space. This is evidenced by the median position error being substantially lower than the average position error, indicating that our pipeline can efficiently and robustly retarget large-scale motion data. Furthermore, the strong performance on the test set, corroborated by the loss curves in Fig.~\ref{fig:main_loss_gr1_h1}, demonstrates that the network generalizes well to unseen motions rather than merely overfitting to the training data.

\begin{figure}[htbp]
    \centering
    \includegraphics[width=0.95\linewidth]{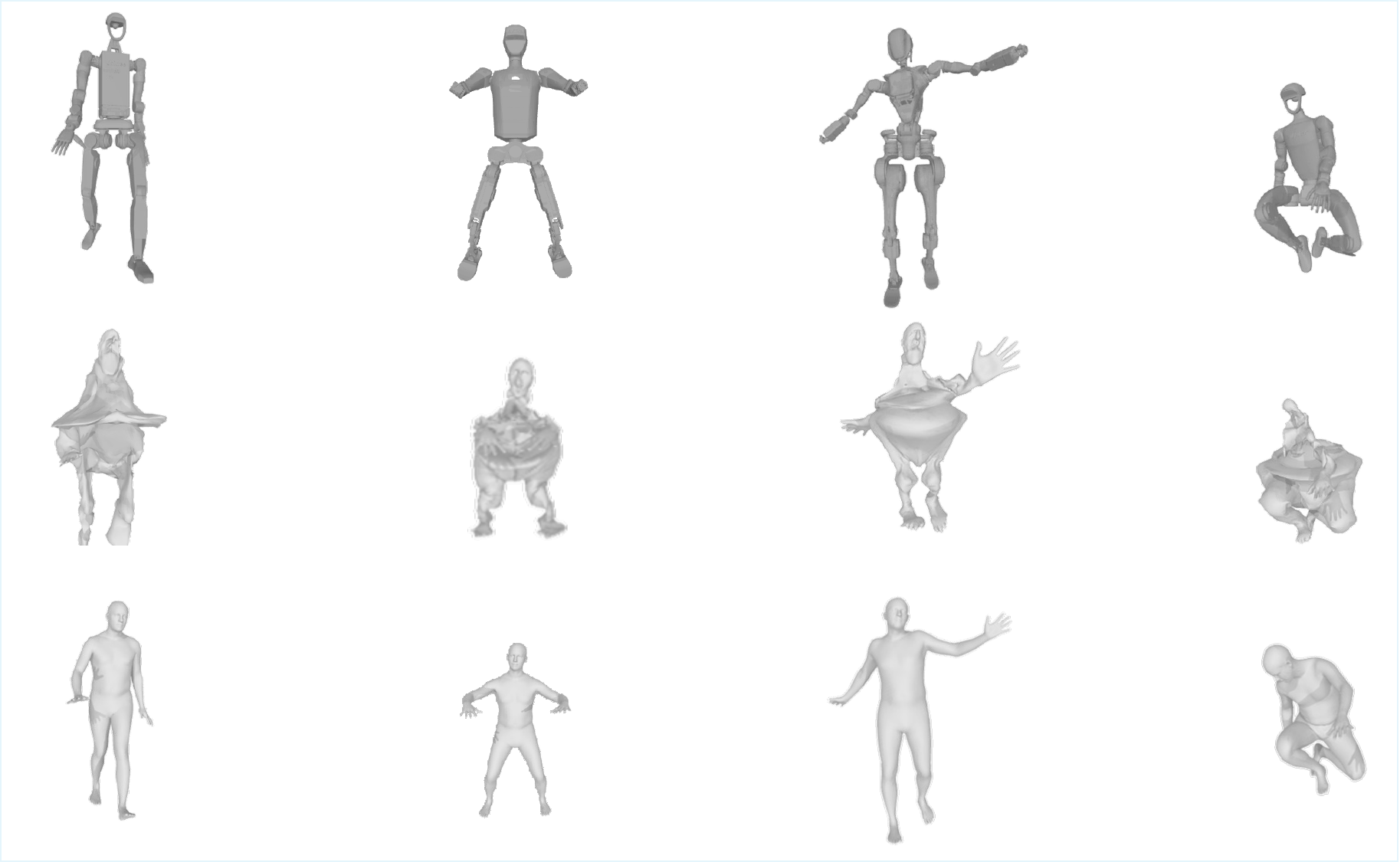}
    \caption{Qualitative retargeting results across different humanoids and motions. From top to bottom: the robot's executed pose, the fitted SMPL+H pose, and the original human reference pose.}
    \label{fig:example_retargeting}
\end{figure}

To provide a qualitative validation of our pipeline, Fig.~\ref{fig:example_retargeting} visually demonstrates the retargeting results for various motions across the four distinct humanoid morphologies. In each set, the final robot motion is shown alongside the fitted SMPL+H model and the original human reference motion, illustrating the high fidelity of the kinematic mapping.

\begin{figure}[htbp]
    \centering
    \includegraphics[width=0.95\linewidth]{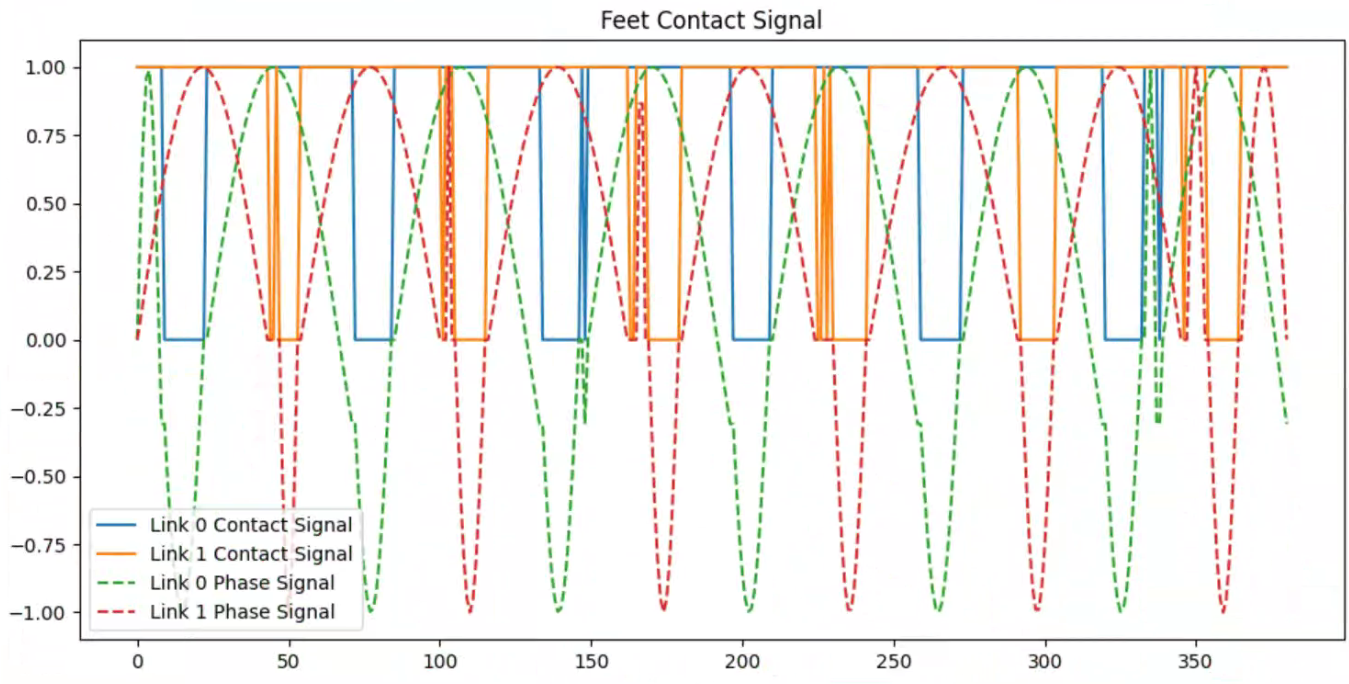}
    \caption{The conversion of a binary foot-contact signal (top) into a continuous sine-cosine phase representation (bottom), which serves as a differentiable and consistent input for the policy network.}
    \label{fig:feet_contact_filtering}
\end{figure}

However, despite the high accuracy of the retargeting network, the raw output can exhibit minor jitter due to the discrete nature of sampling from the source data. As previously shown in Fig.~\ref{fig:filter_and_cyclic_subseq}, a post-processing smoothing filter effectively mitigates this. To further enrich the expert data for learning robust gaits, our pipeline also infers foot-contact events. By analyzing foot velocity and height derived from forward kinematics, we estimate ground contact instances. This binary signal is then converted into a continuous sine-cosine phase sequence, ensuring differentiability and consistency with other RL observations. As depicted in Fig.~\ref{fig:feet_contact_filtering}, the phase at transition points can be fine-tuned using the same environmental parameters as the RL agent. This inferred contact signal achieves a high consistency of 94.8\% with the ground-truth contacts in Isaac Sim (reaching 98.2\% on specific locomotion sub-tasks like AMASS-ACCAD), confirming its suitability as accurate demonstration data.

\begin{figure*}[htbp]
    \centering
    \includegraphics[width=0.9\textwidth]{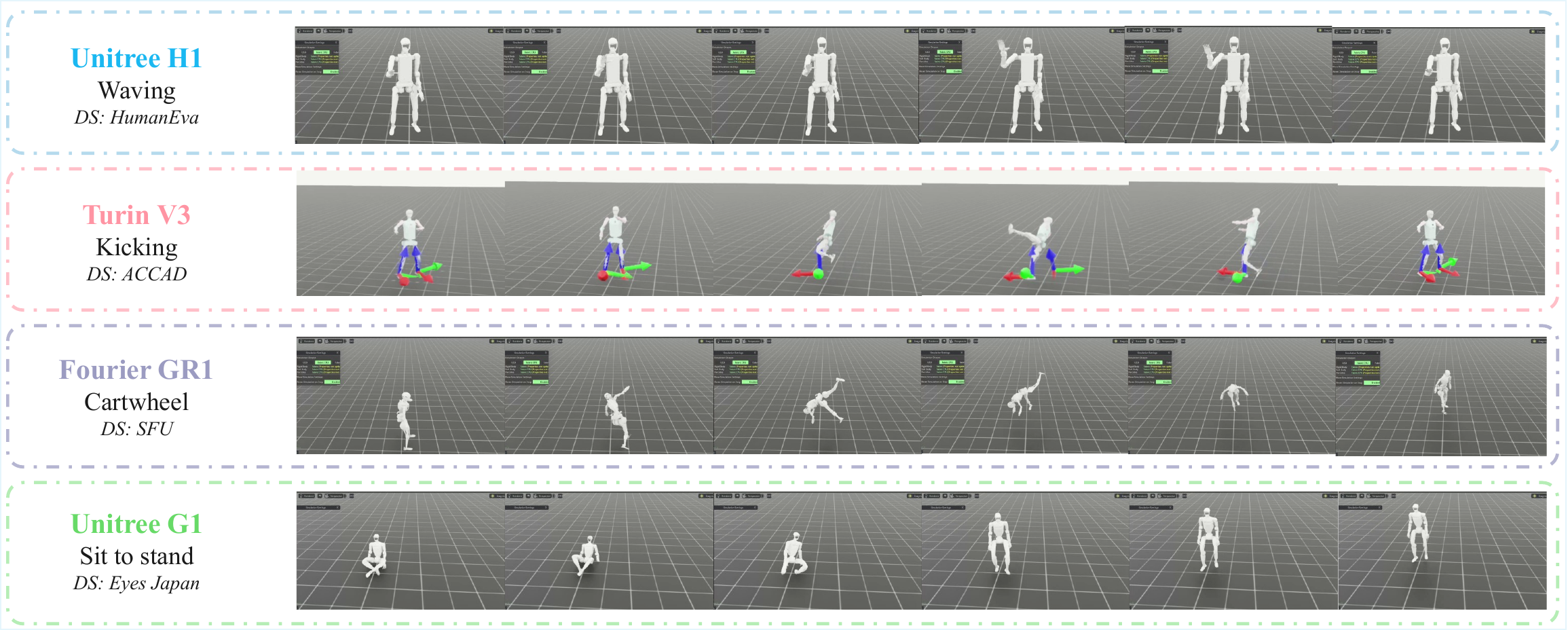}
    \caption{Simulation of the retargeted motions in Isaac Sim, demonstrating the kinematic and dynamic feasibility of the final processed motion data for downstream imitation learning.}
    \label{fig:all_fours_simulation}
\end{figure*}

Finally, Fig.~\ref{fig:all_fours_simulation} presents the final simulation results of our retargeted motions within Isaac Sim. The experiments show that the generated motions are smooth, kinematically coherent, and dynamically plausible, rendering them highly suitable for subsequent imitation learning tasks.

\subsection{MMTransformer Backbone Analysis}
With a reliable motion retargeting pipeline established, we now turn to validating the convergence capabilities of our novel PPO backbone. To this end, we conduct a comparative analysis in two distinct scenarios: a standard RL task focused on command-following, and an IL task centered on tracking a specific reference motion. Given that our framework supports multiple MMTransformer architectures and embedding strategies, we evaluate their efficacy under the configurations detailed in Table~\ref{tab:backbone_exp_setup}.

\begin{table*}[t]
\centering
\caption{Experimental Setup for Backbone Performance Analysis}
\label{tab:backbone_exp_setup}
\begin{tabular}{@{}ll@{}}
\toprule
\textbf{Component} & \textbf{Specification} \\
\midrule
\textbf{Tasks} & 
\begin{tabular}[t]{@{}p{0.85\linewidth}@{}}
\textbf{1. Reinforcement Learning (RL):} Locomotion task based on the official Isaac Lab environment for the Unitree H1, adapted for our H1-2 model configuration. The goal is to follow velocity commands. \\
\textbf{2. Imitation Learning (IL):} Tracking two motions from the `ACCAD/Male1Walking' dataset: (a) walk forward and turn, and (b) walk backwards. The episode length for tracking is set to 20 seconds.\textsuperscript{\textdagger}
\end{tabular}
\\ \addlinespace 
\textbf{Models Under Test} &
\begin{tabular}[t]{@{}p{0.85\linewidth}@{}}
\textbf{1. MLP (Baseline):} Standard Multi-Layer Perceptron with hidden layers [512, 256, 128]. For imitation part, AMP is also enabled.\\
\textbf{2. MMTransformer + Obs. Embedding:} Our transformer backbone with the basic embedding layer. \\
\textbf{3. MMTransformer + Hybrid Strategy:} A weighted combination of the MMTransformer and MLP policy outputs. \\
\textbf{4. MMTransformer + Obs. Embedding V2:} Our transformer with the advanced grouped embedding layer.\textsuperscript{\textdaggerdbl}
\end{tabular}
\\ \addlinespace
\textbf{Observation Space} &
\begin{tabular}[t]{@{}p{0.85\linewidth}@{}}
\textbf{RL Observations:} Linear \& angular velocity, projected gravity, joint positions \& velocities, foot-contact phase. \\
\textbf{IL Reference Observations:} Reference linear \& angular velocity, ref. projected gravity, ref. joint positions \& velocities, ref. link positions \& velocities, ref. height, ref. foot-contact phase.
\end{tabular}
\\ \addlinespace
\textbf{Hyperparameters} & 
\begin{tabular}[t]{@{}p{0.85\linewidth}@{}}
\textbf{Transformer:} $d_{\text{model}}=256$, $N_{\text{layers}}=2$, $N_{\text{heads}}=8$. \\
\textbf{Training:} A fixed random seed of 42 is used to ensure consistency in environment randomization.
\end{tabular}
\\
\bottomrule
\multicolumn{2}{p{0.95\linewidth}}{\textsuperscript{\textdagger}\footnotesize{While our data processing allows for indefinitely long motion sequences, we cap the episode length at 20s to ensure efficient training cycles.}} \\
\multicolumn{2}{p{0.95\linewidth}}{\textsuperscript{\textdaggerdbl}\footnotesize{In this V2 configuration, observations are grouped by modality: (a) IMU-related values (angular velocity, gravity), (b) foot-contact phases, (c) all joint states, and (d) other observations in separate groups.}}
\end{tabular}
\end{table*}

To quantify the performance in these experiments, we define the following metrics:
\begin{itemize}
    \item \textbf{For the RL Task:} We measure the policy's ability to achieve stable and proficient locomotion.
    \begin{itemize}
        \item \textit{Stable Epoch:} The training epoch after which the average episode length consistently exceeds 90\% of the maximum episode duration. This indicates that the robot has learned to avoid falling.\textsuperscript{1}
        \item \textit{Well-Trained Epoch:} The epoch where the reward for tracking velocity commands is consistently high. We define this as the point where the velocity tracking reward, $R_{\text{vel}}$, satisfies $R_{\text{vel}} > 0.75$. The reward is calculated as:
        $$
        R_{\text{vel}} = \exp\left(-\frac{\|\mathbf{v} - \mathbf{v}_{\text{target}}\|^2}{2\sigma^2}\right), \quad \text{with } \sigma=0.35
        $$
    \end{itemize}
    \item \textbf{For the IL Task:} In addition to the above, we assess the fidelity of motion tracking.
    \begin{itemize}
        \item \textit{Well-Tracked Epoch:} The epoch where the joint position tracking reward, $R_{\text{pos}}$, consistently exceeds 0.7. The reward function follows a similar exponential form with $\sigma=0.4$.
        \item \textit{Action Similarity Score:} A score calculated at 10,000 training iterations that measures how closely the learned policy's motion resembles the expert demonstration. This score is derived from a pre-trained AMP-style discriminator (which is not used for policy training).\textsuperscript{2} The score is defined as:
        $$
        S_{\text{similarity}} = \frac{s_{\text{action}} - \bar{s}_{\text{neg}}}{s_{\text{pos}} - s_{\text{action}}}
        $$
        where $s_{\text{action}}$ is the discriminator's score for the generated motion sequence, $\bar{s}_{\text{neg}}$ is the average score for non-expert data, and $s_{\text{pos}}$ is the score for the expert data. A higher score indicates that the learned motion is more successful at "fooling" the discriminator, implying a closer resemblance to the reference motion.
    \end{itemize}
\end{itemize}
\footnotetext[1]{The episode length tracks the time until the environment is reset. When this duration approaches the maximum allowed time, it implies the agent rarely fails (e.g., falls) and can thus be considered stable.}
\footnotetext[2]{We train a discriminator on agent observations vs. expert observations but do not use its output as a reward signal for the policy, only for this final evaluation metric.}

The performance of these model architectures on the RL and IL tasks is summarized in Table~\ref{tab:backbone_performance_results}. The results provide clear insights into the strengths and weaknesses of each approach.

\begin{table*}[t]
\centering
\begin{threeparttable}
        \caption{Performance Comparison of Backbone Architectures on RL and IL Tasks} 
        \label{tab:backbone_performance_results} 
        \begin{tabular}{@{}l S[table-format=3] c c c c c@{}}
            \toprule
            {\textbf{Model Architecture}} & \multicolumn{2}{c}{\textbf{RL Task}} & \multicolumn{4}{c}{\textbf{IL Task}} \\
            \cmidrule(lr){2-3} \cmidrule(lr){4-7}
             & {Stable} & {Well-Trained} & {Stable} & {Well-Trained} & {Well-Tracked} & {Similarity} \\
             & {Epoch} & {Epoch} & {Epoch} & {Epoch} & {Epoch} & {Score} \\
            \midrule
            1. MLP (Baseline)\tnote{a} & 120 & 980 & {$\infty$}(2640) & {$\infty$}($>1e4$) & {$\infty$}($>1e4$) & -0.42(-0.11) \\
            2. MMT + Basic Emb. & 290 & 1400 & 560 & {$\infty$} & 4490 & 0.24 \\
            3. MMT-MLP Hybrid & \textbf{87} & \textbf{650} & 380 & 3420 & 3750 & 0.43 \\
            4. MMT + Grouped Emb. (V2) & 95 & 720 & \textbf{160} & \textbf{1870} & \textbf{2510} & \textbf{0.89} \\
            \bottomrule
        \end{tabular}
        \begin{tablenotes}
            \item[a] \footnotesize For testing MLP adapted for IL tasks, value inside brackets are the test result with AMP enabled.
        \end{tablenotes}
    \end{threeparttable}
\end{table*}

Our analysis of these results is as follows:
\begin{itemize}
    \item \textbf{MLP (Baseline):} The standard MLP architecture is proficient at the basic command-following RL task, achieving stability and proficiency within a reasonable number of epochs. However, it completely fails on the imitation learning task. Its simple structure is incapable of effectively processing the high-dimensional, multi-modal reference observations required for motion tracking. As a result, the policy never learns to walk stably, leading to an Action Similarity Score that is even lower than the discriminator's average score for non-expert data. Even with the assistance of AMP to guide the correct convergent path, the structure still performs poorly compared to other MMTransformer based structures.

    \item \textbf{MMT + Basic Emb.:} The basic MMTransformer architecture, while structurally more suitable for multi-modal fusion, exhibits slower convergence on the pure RL task. This is likely due to a larger parameter space and a naive embedding strategy that flattens observations without preserving their inherent structure, making optimization more challenging. Conversely, its ability to model the relationship between agent and reference states allows it to successfully tackle the IL task, where the MLP failed.

    \item \textbf{MMT-MLP Hybrid:} This hybrid model, which combines the outputs of both networks through a learnable weighted fusion, demonstrates impressive performance on \textit{both} the RL and IL tasks when trained individually. It leverages the strengths of both architectures. However, its critical limitation is the MLP component's inability to handle masked inputs, which is the core mechanism for switching between reference-based (IL) and reference-free (RL) control within a single policy. Therefore, this architecture is unsuitable for a unified learning framework.

    \item \textbf{MMT + Grouped Emb. (V2):} Our final proposed architecture effectively addresses the shortcomings of the other models. By introducing an inductive bias through modality-based grouping and convolutional pre-processing of observations, the Grouped Embedding (V2) preserves crucial structural and temporal information that is lost in the basic embedding. This leads to performance that surpasses the MLP baseline on the RL task. More importantly, it significantly outperforms all other architectures on the IL task, demonstrating the most rapid and effective learning. This model's superior ability to process, fuse, and act upon complex observational data confirms its suitability for training sophisticated and unified imitation learning policies.
\end{itemize}

\subsection{DAgger-MMPPO Algorithm Evaluation}
To validate that our GBC framework can train a universal controller, we must verify the effectiveness of the DAgger-MMPPO algorithm design. This section focuses on evaluating the contributions of the DAgger network for multi-action tracking and the Toddler Curriculum for improving stability. To rigorously test our algorithm under challenging conditions, we selected the \textbf{Unitree H1-2} as the robotic platform for this analysis. Among our four tested configurations, the H1-2 possesses the greatest mass and height, making its training convergence significantly more difficult and thus better highlighting the benefits of our algorithmic designs. The detailed experimental setup is outlined in Table~\ref{tab:dagger_mppo_exp_setup}.

\begin{table*}[t]
\centering
\caption{Experimental Setup for DAgger-MMPPO Algorithm Evaluation}
\label{tab:dagger_mppo_exp_setup}
\begin{tabular}{@{}ll@{}}
\toprule
\textbf{Component} & \textbf{Specification} \\
\midrule
\textbf{Robotic Platform} & \textbf{Unitree H1-2}. Chosen for its challenging dynamics due to high mass and inertia. \\
\addlinespace
\textbf{Tasks} & Three categories of increasing difficulty, all sourced from the AMASS dataset: \\
& \quad \textbf{1. Simple Task:} Tracking basic walking motions from `ACCAD/Male1Walking\_c3d` and `ACCAD/Male2Walking\_c3d`. \\
& \quad \textbf{2. Medium Task:} Tracking a diverse set of randomly sampled motions from the entire `ACCAD` dataset. \\
& \quad \textbf{3. Hard Task:} Tracking complex motions involving rotation and jumping from the `ACCAD/Martial Arts` directory. \\
\addlinespace
\textbf{Curriculum Learning} & A 5-level curriculum is applied to key tracking rewards by annealing the standard deviation $\sigma$: \\
& \quad \textbf{Linear Velocity Reward $\sigma$ Levels:} [0.5, 0.45, 0.42, 0.4, 0.35] \\
& \quad \textbf{Angular Velocity Reward $\sigma$ Levels:} [0.5, 0.45, 0.42, 0.4, 0.35] \\
& \quad \textbf{Joint Position Reward $\sigma$ Levels:} [0.75, 0.6, 0.5, 0.45, 0.4] \\
& \quad \textbf{Update Condition:} The curriculum advances to the next level when the corresponding reward term exceeds 0.7.\textsuperscript{\textdagger} \\
\addlinespace
\textbf{Training Protocol} & Total of \textbf{50,000} training iterations for each run. \\
& For configurations with DAgger, the first \textbf{5,000} iterations are dedicated to pre-training the DAgger network. \\
\addlinespace
\textbf{Evaluation Metrics} & \textbf{1. Stable Epoch:} The epoch where the agent consistently avoids falling. \\
& \textbf{2. Success Rate:} The percentage of successfully converged motions out of all loadable motions in the task set.\textsuperscript{\textdaggerdbl} \\
& \textbf{3. Final Curriculum Level:} The terminal curriculum level achieved for each reward term at the end of training. \\
\addlinespace
\textbf{Algorithm Configs.} & We compare four variations of our algorithm: \\
& \quad \textbf{1. MMPPO (Baseline):} Our policy network trained with PPO but without DAgger or the Toddler Curriculum. \\
& \quad \textbf{2. MMPPO + DAgger:} The baseline augmented with the DAgger training scheme. \\
& \quad \textbf{3. MMPPO + Toddler Curriculum:} The baseline augmented with physics-based assistance. The max assistive force is \\
& \quad \quad set to 730N (equivalent to the robot's weight), and the assistance is gradually annealed as the policy stabilizes. \\
& \quad \textbf{4. MMPPO (Full):} The complete algorithm including both DAgger and the Toddler Curriculum. \\
\bottomrule
\multicolumn{2}{p{0.95\linewidth}}{\textsuperscript{\textdagger}\footnotesize{The update threshold is set to 0.7 to maintain consistency with the "Well-Trained" metric in the previous subsection. In practice, a lower threshold (e.g., 0.5) can be used to create a smoother reward curve.}} \\
\multicolumn{2}{p{0.95\linewidth}}{\textsuperscript{\textdaggerdbl}\footnotesize{Success is defined by the policy's ability to learn a stable tracking behavior for a given motion sequence without persistent failure.}}
\end{tabular}
\end{table*}

The comparative results of these algorithm configurations across the three task categories are presented in Table~\ref{tab:dagger_mppo_results}.

\begin{table*}[t]
\centering
\caption{Performance Evaluation of DAgger-MMPPO Algorithm Configurations Across Tasks of Varying Difficulty}
\label{tab:dagger_mppo_results}
\sisetup{table-format=4.0, table-number-alignment=center, tight-spacing=true} 
\begin{tabular}{@{}l S[table-format=4] S[table-format=2.1, table-space-text-post=\%] c
                  S[table-format=3] S[table-format=2.1, table-space-text-post=\%] c
                  S[table-format=4] S[table-format=2.1, table-space-text-post=\%] c @{}}
\toprule
{\textbf{Algorithm Configuration}} & \multicolumn{3}{c}{\textbf{Simple Task}} & \multicolumn{3}{c}{\textbf{Medium Task}} & \multicolumn{3}{c}{\textbf{Hard Task}} \\
\cmidrule(lr){2-4} \cmidrule(lr){5-7} \cmidrule(lr){8-10}
 & {Stable} & {Succ.} & {Final Curr.} & {Stable} & {Succ.} & {Final Curr.} & {Stable} & {Succ.} & {Final Curr.} \\
 & {Epoch} & {Rate} & {Level\textsuperscript{a}} & {Epoch} & {Rate} & {Level\textsuperscript{a}} & {Epoch} & {Rate} & {Level\textsuperscript{a}} \\
\midrule
MMPPO (Baseline) & 160 & 78.9\% & 3/2/4 & 430 & 82.5\% & 2/1/3 & 1300 & 61.2\% & 1/1/2 \\
MMPPO + DAgger & 230 & 87.9\% & 3/3/5 & 560 & 87.7\% & 3/2/4 & 920 & 73.1\% & 2/1/4 \\
MMPPO + Toddler Curr. & 140 & 90.9\% & 3/4/5 & 420 & 87.0\% & 2/3/5 & 740 & 82.7\% & 1/2/4 \\
MMPPO (Full) & 200 & 90.9\% & 3/4/5 & 500 & 91.3\% & 3/3/5 & 810 & 82.7\% & 2/3/5 \\
\bottomrule
\multicolumn{10}{l}{\textsuperscript{a}\footnotesize{Curriculum levels are displayed in the order of: Linear Velocity / Angular Velocity / Joint Position.}}
\end{tabular}
\end{table*}

It is important to note that the AMASS dataset contains motions that are kinematically challenging or violate the termination conditions for the H1-2 robot's configuration (e.g., crawling, cartwheels, flying kicks). While our framework can learn to execute versions of these motions by tracking key quantities like velocity and joint angles rather than absolute spatial coordinates, the resulting behavior may differ from the reference and not achieve perfect tracking. Since these motions are included in the success rate calculation, no algorithm configuration reaches 100\% success.

From the results, we can observe that on the \textbf{Simple Task}, all four configurations perform effectively with minor differences, though the baseline MMPPO learns more slowly. When introducing \textbf{DAgger}, the policy requires a longer time to reach a stable state. This is because the initial policy distribution $\pi_D$ from the DAgger pre-training does not perfectly align with the target distribution $\bar{\pi}_R$ of the full-physics environment, and the smaller exploration radius $\sigma$ of $\pi_D$ necessitates more training to adapt. However, once stabilized, its convergence is markedly faster than the baseline. The \textbf{Toddler Curriculum} enables the policy to achieve stability much more rapidly; the agent only needs to avoid immediate instability before the assistive force engages, after which it is much less likely to fall. During this assisted, "suspended" phase—which is analogous to the DAgger training environment—the agent can more effectively track joint motions. A side effect is that because the assistive force often acts when the agent is already tilted, the linear velocity tracking can be slightly compromised, while angular velocity tracking becomes more accurate.

\begin{figure*}[htbp]
    \centering
    \includegraphics[width=\textwidth]{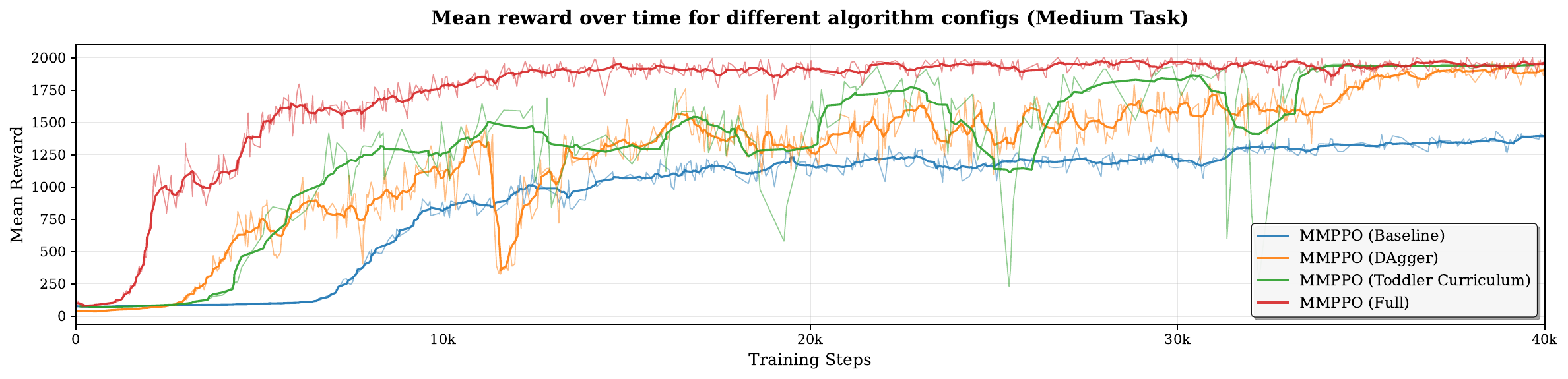}
    \caption{Average reward curves for the four algorithm configurations on the Medium Task. The full MMPPO algorithm (red) demonstrates the fastest and most stable convergence. The Baseline MMPPO (blue) shows gradual improvement but at a much slower rate. The policy with the Toddler Curriculum (green) converges rapidly within each stage but exhibits noticeable performance dips when the curriculum difficulty increases. The DAgger-augmented policy (orange) takes longer to stabilize initially but then shows significantly faster and more stable learning than the baseline.}
    \label{fig:medium_task_reward}
\end{figure*}

The reward curves in Fig.~\ref{fig:medium_task_reward} for the \textbf{Medium Task} illustrate these dynamics. The policy with the Toddler Curriculum converges quickly within each curriculum stage, but experiences a performance drop each time the assistive force is reduced, requiring the policy to re-adapt. This process mirrors the transition from $\pi_D$ to $\pi_R$ in DAgger, but occurs in smaller, more manageable steps. In contrast, the DAgger-augmented policy, while taking longer to stabilize initially, exhibits a much steeper and more stable learning curve than the baseline. Finally, our \textbf{full MMPPO algorithm} synergizes these benefits: the DAgger component enhances training efficiency and stability, while the Toddler Curriculum ensures a rapid and structured progression through its staged difficulty, resulting in the most robust and highest-performing policy.

\subsection{Generalization and Transferability Assessment}
In this final experimental section, we evaluate the generalization and transfer learning capabilities of policies trained with the GBC framework. We first present the domain randomization settings used during training to enhance policy robustness. Subsequently, we validate the following key characteristics of the trained policies:
\begin{enumerate}
    \item The ability of a policy trained on a \textbf{single motion type} to imitate out-of-distribution (O.O.D.) motions.
    \item The ability of a policy trained on \textbf{multiple motion types} to generalize to O.O.D. motions.
    \item The performance of a unified policy when its reference observation input is disabled (i.e., \texttt{ref\_observation} is \texttt{None} and its corresponding tokens are masked, as shown in Fig.~\ref{fig:mmtransformer}), forcing it to rely solely on agent-centric observations to follow velocity commands.
\end{enumerate}

To foster robust and transferable policies, we employ a comprehensive suite of domain randomization techniques during training. These include varying dynamics parameters (e.g., payload mass by $\pm 5.0$~kg; link masses, joint stiffness, and damping by multipliers in $[0.8, 1.2]$), randomizing each episode's initial state to match a random point in the reference motion, and applying external pushes of $\pm 0.5$~m/s to the robot's base at random intervals. Apart from that, we add uniformly distributed noises to observation and reference observation terms to simulate sensor noises and poor demonstration inputs, in order to increase the robustness of the trained policy.

Our experimental protocol is designed as follows: We first train two separate policies using motions from the ACCAD sub-dataset: one on the \textbf{Walking Task} (imitating only walking motions) and another on the \textbf{Medium Task} (imitating randomly sampled motions, as in the previous section). We then compare their average tracking error on both in-distribution test data and on O.O.D. trajectories generated from our own motion recordings. After identifying the better-performing policy from this comparison, we evaluate its command-following locomotion capabilities in Isaac Sim with the reference observation input disabled. Finally, we assess the policy's sim-to-sim transferability by deploying it in the MuJoCo simulator, thereby demonstrating the potential of our framework for producing deployable policies.

\begin{figure}[htbp]
    \centering
    \includegraphics[width=\linewidth]{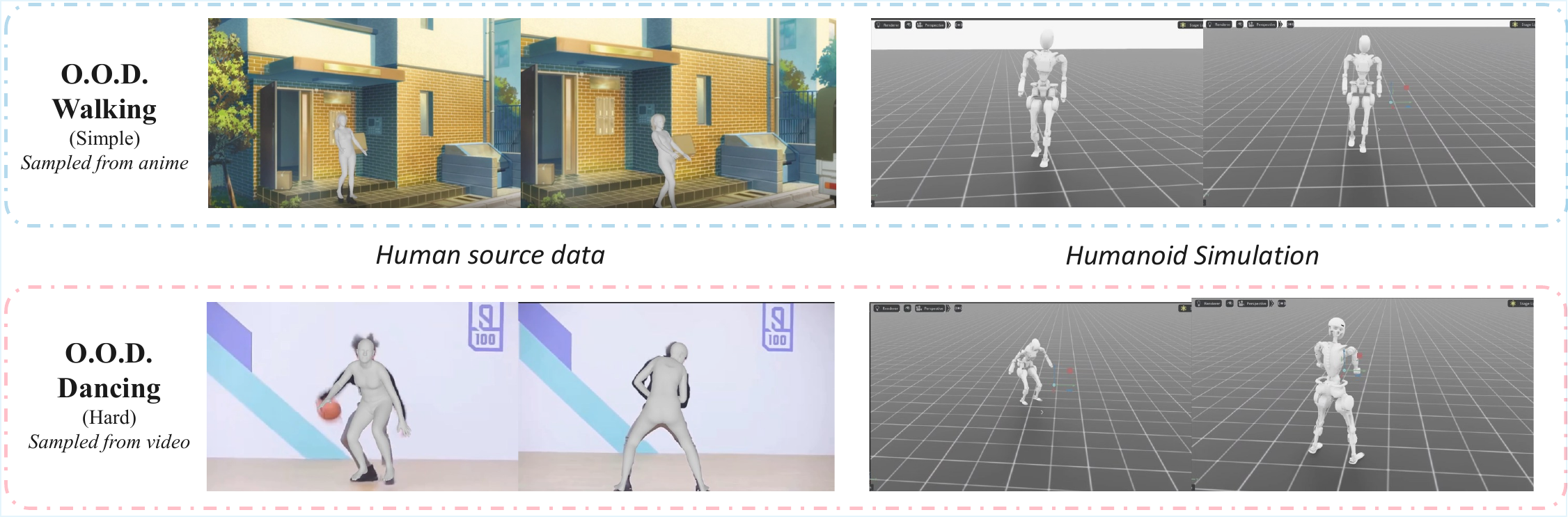}
    \caption{The Out-of-Distribution (O.O.D.) reference motions and their corresponding execution in the Isaac Sim environment.}
    \label{fig:ood_data}
\end{figure}

The O.O.D. data used for our assessment includes a basic walking motion (\textit{O.O.D. Walking}) and a complex dance sequence (\textit{O.O.D. Dancing}), as shown in Fig.~\ref{fig:ood_data}. These motions were captured using WHAM~\cite{shin2024wham} and processed through our data pipeline to generate executable actions. To evaluate the tracking quality, we report the average tracking index, defined as $\exp(-\|\text{err}\|^2/\sigma^2)$, after 10 seconds of tracking. This index is calculated for joint positions\footnote{For the multi-joint robot, the error is the L2 norm of the joint position differences, reflecting the largest joint angle deviations during tracking.}, linear velocity, and angular velocity, using $\sigma=0.4$ to characterize the distribution of the policy's tracking state. To specifically validate the policy's ability to track complex joint motions, the Fourier GR1 robot is used for this set of experiments.

Table~\ref{tab:ood_performance} presents the performance of the checkpoints trained on the Walking Task and the Medium Task, evaluated on both in-distribution and O.O.D. data.

\newcolumntype{C}{>{\centering\arraybackslash}X}

\begin{table*}[t]
\centering
\caption{TRACKING PERFORMANCE ON IN-DISTRIBUTION AND OUT-OF-DISTRIBUTION DATA}
\label{tab:ood_performance}
\begin{tabularx}{\textwidth}{@{}l CCCCCC@{}}
\toprule
{\textbf{Data Input}} & \multicolumn{3}{c}{\textbf{Walking Task Ckpt.}} & \multicolumn{3}{c}{\textbf{Medium Task Ckpt.}} \\
\cmidrule(lr){2-4} \cmidrule(lr){5-7}
 & Joint Idx. & Lin. Vel. Idx. & Ang. Vel. Idx. & Joint Idx. & Lin. Vel. Idx. & Ang. Vel. Idx. \\
\midrule
Training Distribution & 0.867 & 0.741 & 0.656 & 0.824 & 0.682 & 0.610 \\
O.O.D. Walking & 0.833 & 0.702 & 0.845 & 0.817 & 0.688 & 0.801 \\
O.O.D. Dancing & 0.282 & 0.329 & 0.364 & 0.724 & 0.597 & 0.633 \\
\bottomrule
\multicolumn{7}{p{0.97\textwidth}}{\footnotesize{\textbf{Note:} The directly converted reference motions are not always dynamically feasible. Our RL policy learns to imitate the closest dynamically viable motion, thus the tracking index does not reach 1.0. A higher index still indicates a closer resemblance to the original motion. The linear and angular velocities from WHAM are estimated via SLAM and contain inherent inaccuracies.}}
\end{tabularx}
\end{table*}

The results indicate that the policy trained solely on walking data possesses some transferability, enabling it to track a novel walking motion that is similar to its training distribution. However, it fails to generalize to the drastically different O.O.D. dancing motion. In contrast, the policy trained on the more diverse Medium Task learns a richer set of motor primitives. While its tracking fidelity on specific tasks may be slightly lower than that of a specialized model, its generalization performance is far superior, allowing it to successfully track and execute the novel dancing motion. This also highlights that when there is a mismatch between the reference velocity and the robot's capabilities, the policy prioritizes mimicking joint positions and foot-contact patterns to maintain physical stability, a behavior shaped by our reward function design.

Furthermore, we evaluated the performance of the trained policies with their reference inputs disabled. Both the Walking Task and Medium Task policies were able to follow velocity commands for basic locomotion. Due to the limited amount of walking data in the training set, the policies could consistently track the direction of the velocity command, but the precision of the speed tracking could be improved. Notably, when commanded with a lateral velocity, the Medium Task policy attempts to perform a side-leap (a motion present in its training data), whereas the Walking Task policy does not.

\begin{figure*}[htbp]
    \centering
    \includegraphics[width=0.92\textwidth]{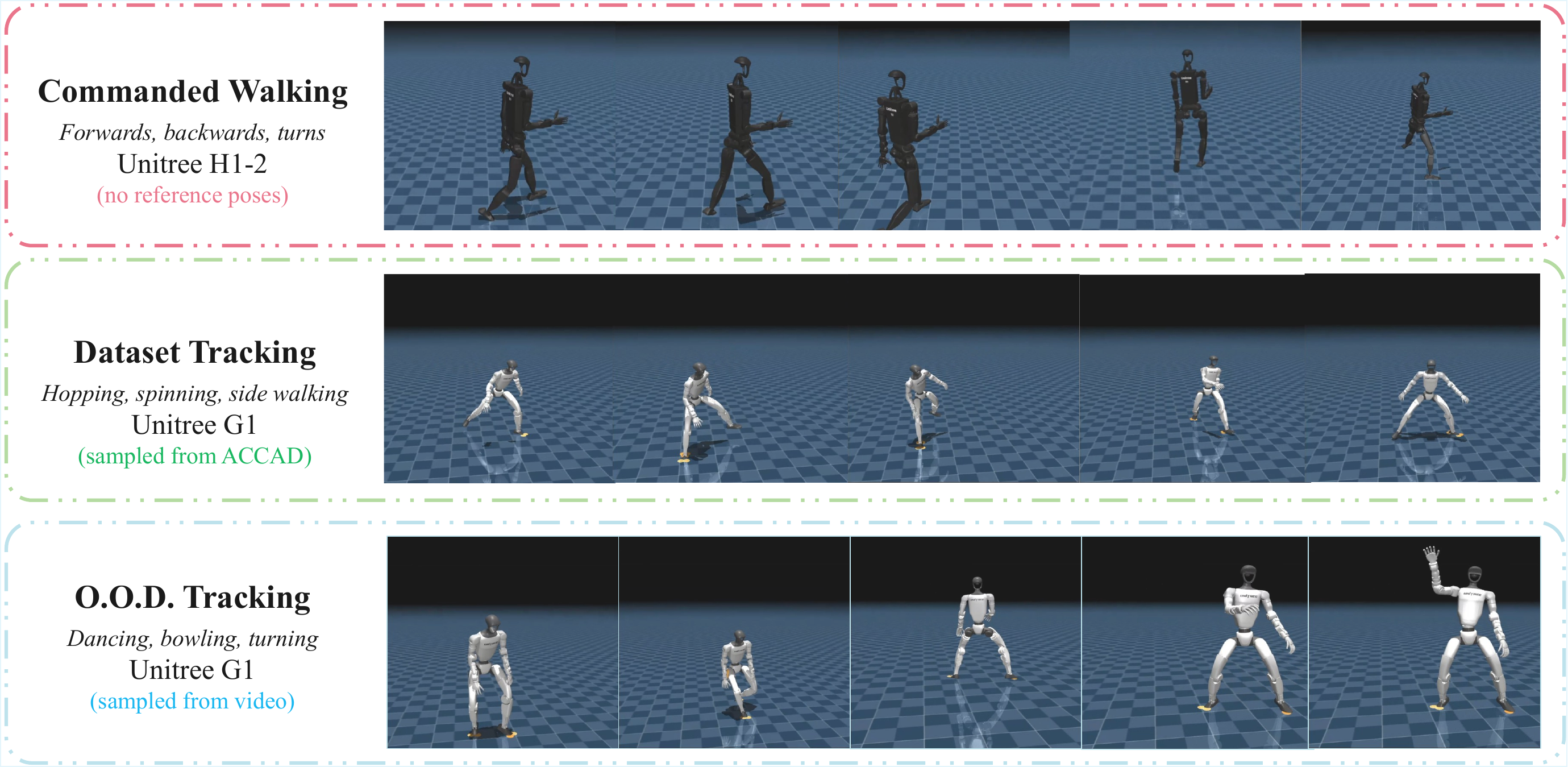}
    \caption{Simulation results of the trained policies deployed in the MuJoCo platform. The first column (H1-2) shows the policy trained on the Walking Task executing command-following behaviors (forward, backward, turning) with reference inputs disabled. The second column (G1) demonstrates the imitation capability of the Medium Task policy as it tracks a motion sampled from its training distribution. The third column (G1) showcases the policy's generalization to O.O.D. data; it can still perform the tracking task, although foot contact and some joint tracking are less precise compared to its performance on in-distribution reference motions.}
    \label{fig:mujoco_sim}
\end{figure*}

Finally, to assess the policy's sim-to-sim transferability, we retrained checkpoints for two specific platforms. The policy trained on the \textbf{Walking Task} was validated on the Unitree H1-2, while the more complex \textbf{Medium Task} policy was validated on the more dexterous Unitree G1. We then selected the G1 checkpoint for deployment in the MuJoCo simulator.\footnote{The Fourier GR1 was not used for this sim-to-sim validation because a reliable MJCF model could not be generated from the provided URDF, even with official conversion tools, making cross-platform testing infeasible.} The results, shown in Fig.~\ref{fig:mujoco_sim}, demonstrate that our policy can effectively control the G1 robot's motion in MuJoCo, successfully performing both reference-based tracking and reference-free command following. This successful transfer confirms that the learned policy is not overfitted to the Isaac Sim environment and, aided by domain randomization, possesses the robustness to handle discrepancies between simulators, indicating strong potential for sim-to-real deployment.


\section{Limitations and Future Work}

Our work, while establishing a comprehensive framework, has several limitations that also define exciting opportunities for future research.

\subsection{Limitations and Mitigations}

\textbf{Sim-to-Real Gap and Physical Deployment.} The most significant limitation of this study is the validation of policies exclusively within simulated environments. Although our extensive domain randomization and sim-to-sim transfer experiments provide strong evidence for policy robustness, the ultimate test of generality lies in real-world deployment. The GBC framework is intentionally designed with sim-to-real transfer in mind, but quantifying its performance on a diverse range of physical humanoid hardware remains the most crucial next step.

\textbf{Optimality of Retargeted Demonstrations.} Our data processing pipeline, while effective, is primarily data-driven and does not incorporate direct feedback from physics-based validation during the retargeting process. Consequently, as observed in our experiments, some reference motions are not fully dynamically feasible for the target robot. Our RL policy learns to find the closest viable motion, but this indicates that the initial demonstration data is not perfect. Future work could improve demonstration quality by integrating physics-in-the-loop validation, as suggested by related works \cite{yagi2024unsupervised, meixner2024towards}, or leveraging existing ground-truth datasets for supervised training, such as the Unitree Retargeting Dataset \cite{lv_2025_lafan1_retargeting}.

\textbf{Hyperparameter Sensitivity.} The DAgger-MMPPO algorithm, a core component of our framework, involves a substantial number of hyperparameters. These include not only the PPO learning parameters but also the annealing schedule for the DAgger intervention coefficient $\rho$, LoRA configurations, and the multi-stage curriculum learning parameters. While our experiments demonstrate a robust configuration, optimizing these hyperparameters for new robot morphologies or vastly different task domains can be a non-trivial endeavor and would benefit from automated tuning methods.

\textbf{Scope of Imitation Tasks.} While the GBC data pipeline is capable of processing hand degrees of freedom, our current experimental validation focuses on whole-body locomotion and gross motor skills. We have not explicitly incorporated the imitation of dexterous hand manipulation. Extending the framework to learn fine-grained, contact-rich manipulation tasks is a clear and valuable direction for future expansion.

\subsection{Future Directions}

The GBC framework serves as a robust foundation for moving beyond pure motion imitation towards more complex, goal-oriented behaviors. Our future work will focus on integrating the learned human-like motor skills with higher-level cognitive abilities.

We are actively exploring the development of generative motion models, such as diffusion models, trained on the large-scale, multi-robot datasets produced by our pipeline. By augmenting these datasets with textual labels, we aim to create policies that can generate novel, human-like motions from natural language commands. Furthermore, we are integrating our framework with Vision Language Models (VLMs) to tackle complex task planning in interactive environments, targeting emerging benchmarks like HumanoidBench \cite{sferrazza2024humanoidbench}.

Ultimately, we envision GBC evolving into a cornerstone platform for the community. Through continued open-source development, we aspire for this framework to empower a broad spectrum of research in humanoid robotics, from enhancing sim-to-real transfer to building the next generation of embodied AI agents capable of both natural movement and intelligent interaction.

\section{Conclusion}
In this paper, we presented the GBC (Generalized Behavior Cloning) framework, a unified approach to the challenging problem of creating generalizable, whole-body controllers for heterogeneous humanoid robots. The GBC framework is designed to overcome the fragmentation of previous methods by integrating the entire pipeline, from universal MoCap data retargeting to robust policy learning. We proposed a novel MMTransformer architecture and a DAgger-MMPPO algorithm, and demonstrated their effectiveness through extensive experiments. Our results show that the framework can successfully train policies on multiple, morphologically distinct humanoids, enabling high-fidelity imitation and effective generalization to novel motions. By releasing GBC as an open-source project, we aim to provide the community with a practical tool to facilitate future research. It is our hope that this work will serve as a valuable contribution towards a future where robots can learn to move with greater versatility and naturalness.




 
%

\bibliographystyle{IEEEtran}

\vfill

\end{document}